\documentclass[10pt,journal,compsoc]{IEEEtran}
\ifCLASSOPTIONcompsoc
  \usepackage[nocompress]{cite}
\else
  \usepackage{cite}
\fi

\ifCLASSINFOpdf
\else
\fi

\usepackage{graphicx}%
\usepackage{multirow}%
\usepackage{amsmath,amssymb,amsfonts}%
\usepackage{amsthm}%
\usepackage{mathrsfs}%
\usepackage{xcolor}%
\usepackage{textcomp}%
\usepackage{manyfoot}%
\usepackage{booktabs}%
\usepackage{algorithmicx}%
\usepackage{algpseudocode}%
\usepackage{listings}%

\usepackage{booktabs}
\usepackage{graphicx}
\usepackage{amsmath}
\usepackage{amssymb}
\usepackage{booktabs}
\usepackage{multirow}
\usepackage{makecell}
\usepackage{amssymb}
\usepackage{bbm}
\usepackage{color}
\usepackage{bbding}
\usepackage{colortbl}
\usepackage{framed}
\usepackage{url}

\newcommand{\keypoint}[1]{\vspace{0.1cm}\noindent\textbf{#1}}
\usepackage[colorlinks=true,linkcolor=magenta,citecolor=green,urlcolor=magenta,]{hyperref}
\usepackage[ruled, vlined, linesnumbered]{algorithm2e}

\begin{document}
\title{Reliable Few-shot Learning under Dual Noises}

\author{Ji~Zhang,
        Jingkuan~Song,
        Lianli~Gao,
        Nicu Sebe, 
        and Heng Tao~Shen,~\IEEEmembership{Fellow~IEEE} 
\IEEEcompsocitemizethanks{
\IEEEcompsocthanksitem Ji Zhang is with the School of Computing and Artificial Intelligence, Southwest Jiaotong University, China. 
\IEEEcompsocthanksitem Lianli Gao is with the School of Computer Science and Engineering, University of Electronic Science and Technology of China, China. 
\IEEEcompsocthanksitem Nicu Sebe is with the Department of Information Engineering and
Computer Science, University of Trento, Italy.
\IEEEcompsocthanksitem Jingkuan Song and Heng Tao Shen are with the School of Computer Science and Technology, Tongji University, China.
\IEEEcompsocthanksitem A preliminary version of this work has been published in ICCV 2023 \cite{Zhang_2023_ICCV}.
}}

\markboth{Journal of \LaTeX\ Class Files,~Vol.~14, No.~8, August~2015}%
{Shell \MakeLowercase{\textit{et al.}}: Bare Advanced Demo of IEEEtran.cls for IEEE Computer Society Journals}
\IEEEtitleabstractindextext{%
\begin{abstract}

Recent advances in model pre-training give rise to task adaptation-based few-shot learning (FSL), 
where the goal is to adapt a pre-trained task-agnostic model for capturing task-specific knowledge with a few-labeled support samples of the target task.
\textcolor{black}{Nevertheless, existing approaches may still fail in the open world due to the inevitable \textit{in-distribution (\textbf{ID})} and \textit{out-of-distribution (\textbf{OOD})} noise from both support and query samples of the target task.}
With limited support samples available, \textbf{i}) the adverse effect of the dual noises can be severely amplified during task adaptation, and \textbf{ii}) the adapted model can produce unreliable predictions on query samples in the presence of the dual noises.
\textcolor{black}{In this work, we propose \textbf{DE}noised \textbf{T}ask \textbf{A}daptation (\textbf{DETA}++) for reliable FSL.} 
\textcolor{black}{DETA++ uses a Contrastive Relevance Aggregation (CoRA) module to calculate image and region weights for support samples, based on which a \textit{clean prototype} loss and a \textit{noise entropy maximization} loss are proposed to achieve noise-robust task adaptation.}
Additionally, DETA++ employs a memory bank to store and refine clean regions for each inner-task class, based on which a Local Nearest Centroid Classifier (LocalNCC) is devised to yield noise-robust predictions on query samples.
Moreover, DETA++ utilizes an Intra-class Region Swapping (IntraSwap) strategy to rectify ID class prototypes during task adaptation, enhancing  the model's robustness to the dual noises.
Extensive experiments demonstrate the effectiveness and flexibility of DETA++.
Code: \url{https://github.com/JimZAI/DETA-plus}.

\end{abstract}

}
\maketitle
\IEEEdisplaynontitleabstractindextext
\IEEEpeerreviewmaketitle

\ifCLASSOPTIONcompsoc
\IEEEraisesectionheading{\section{Introduction}\label{sec:introduction}}
\else
\label{sec:introduction}
\fi

\IEEEPARstart{F}{ew-Shot} Learning (FSL) seeks to rapidly capture new knowledge from a few-labeled support/training samples, a central capability that humans naturally possess, but “data-hungry” machines still lack. 
Over the past years, a community-wide enthusiasm has been ignited to narrow the gap between machine and human intelligence,
especially in the fields of computer vision \cite{snell2017prototypical,he2022masked}, machine translation\cite{xing2019adaptive,bragg2021flex}, reinforcement learning \cite{finn2017model,hong2021reinforced}. 

Generally, the formulation of FSL involves two stages: 
{\textbf{i})} {training-time} task-agnostic knowledge accumulation, and {\textbf{ii})} {test-time} task-specific knowledge acquisition, i.e. task adaptation.
Particularly, the former stage seeks to pre-train a task-agnostic model on large amounts of training samples collected from a set of {base} classes, while the latter targets adapting the pre-trained model for capturing task-specific knowledge of the few-shot (or test) task with {novel} classes, given a tiny set of labeled support samples. 
Early progress in FSL has been predominantly achieved using the idea of meta-learning \cite{zhang2022meta,sun2020meta,ye2022revisiting,ye2022revisiting}, which aligns the objectives of the two stages to better generalize the accumulated knowledge towards few-shot tasks.
Nevertheless, previous studies \cite{chen2019closer,li2021universal,li2022cross} as well as our recent work \cite{luo2023closer} have demonstrated that the two stages can be completely disentangled, and performing test-time task adaptation on any pre-trained models—no matter what training fashions they were learned by, can be more effective than sophisticated meta-learning algorithms.
Furthermore, with the recent success in model pre-training techniques  \cite{he2020momentum,han2021transformer,liu2021swin}, designing effective and efficient {full-finetuning based} \cite{asimplebaseline,hu2022pushing}, {adapter-finetuning based} \cite{li2022cross,xuexploring,li2021universal}, and prompt tuning based \cite{zhu2022prompt,chen2022prompt,jia2022vpt} task adaptation algorithms that can borrow the freely available knowledge from a wide spectrum of pre-trained models is therefore of great practical value, and has demonstrated promise in FSL.

\begin{figure}
\setlength{\abovecaptionskip}{0.1cm}  
		\centering 
		\includegraphics[width=0.79\linewidth]{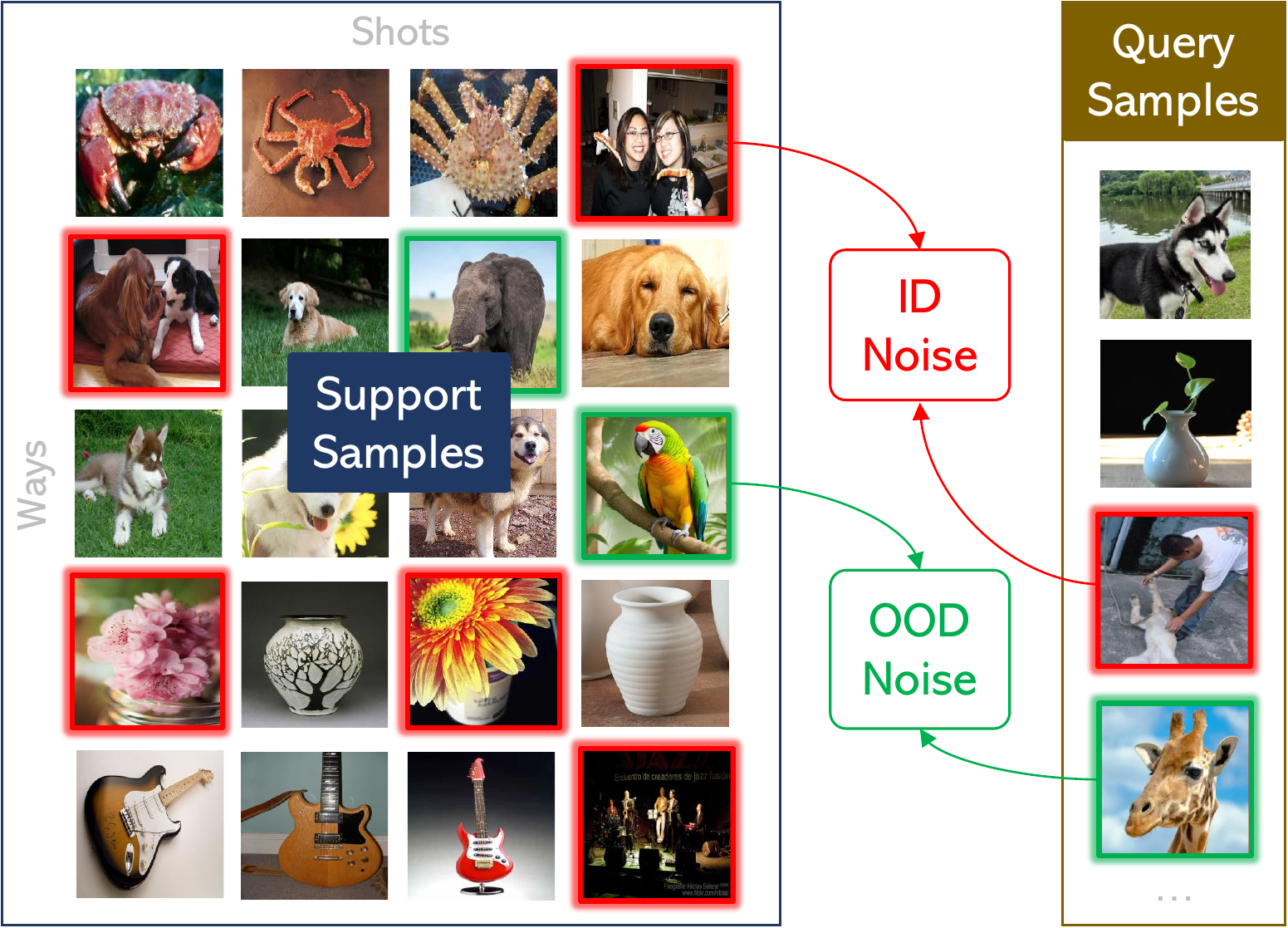}
		\caption{\textcolor{black}{Dual noises in the support (training) set and query (test) set of few-shot tasks. \textbf{i}) \textbf{ID Noise}: in-distribution (ID) samples with cluttered image backgrounds. \textbf{ii}) \textbf{OOD Noise}: out-of-distribution (OOD) samples, i.e., samples from unknown classes.}}
		\label{f1}
\end{figure}

\begin{figure} 
\setlength{\abovecaptionskip}{0.1cm}  
	\centering
	\includegraphics[width=1\linewidth]{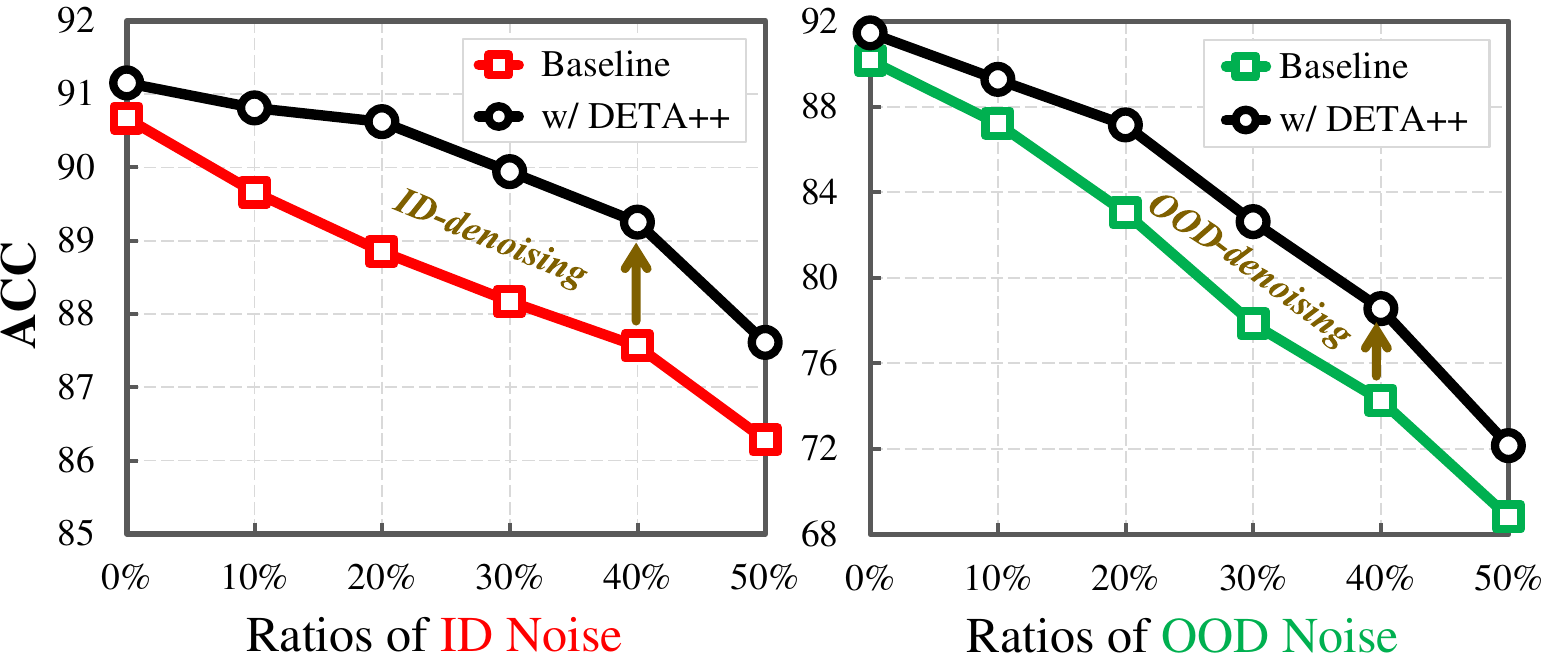}
	\caption{Quantitative evidences that both ID and OOD noise in support samples negatively affect the task adaptation performance. The results are the avg of 100 5-way 10-shot tasks sampled from the 5 classes in Fig. \ref{f1}. ID noise are selected from all samples of the five classes. OOD noise are selected from unseen classes. The baseline scheme is TSA \cite{li2022cross} with a ResNet-18 backbone pre-trained on ImageNet \cite{triantafillou2019meta}.}
	\label{f2}  
\end{figure} 

\begin{figure*}[ht]
\setlength{\abovecaptionskip}{0.1cm}  
\setlength{\belowcaptionskip}{-0.2cm} 
	\centering
	\includegraphics[width=1\linewidth]{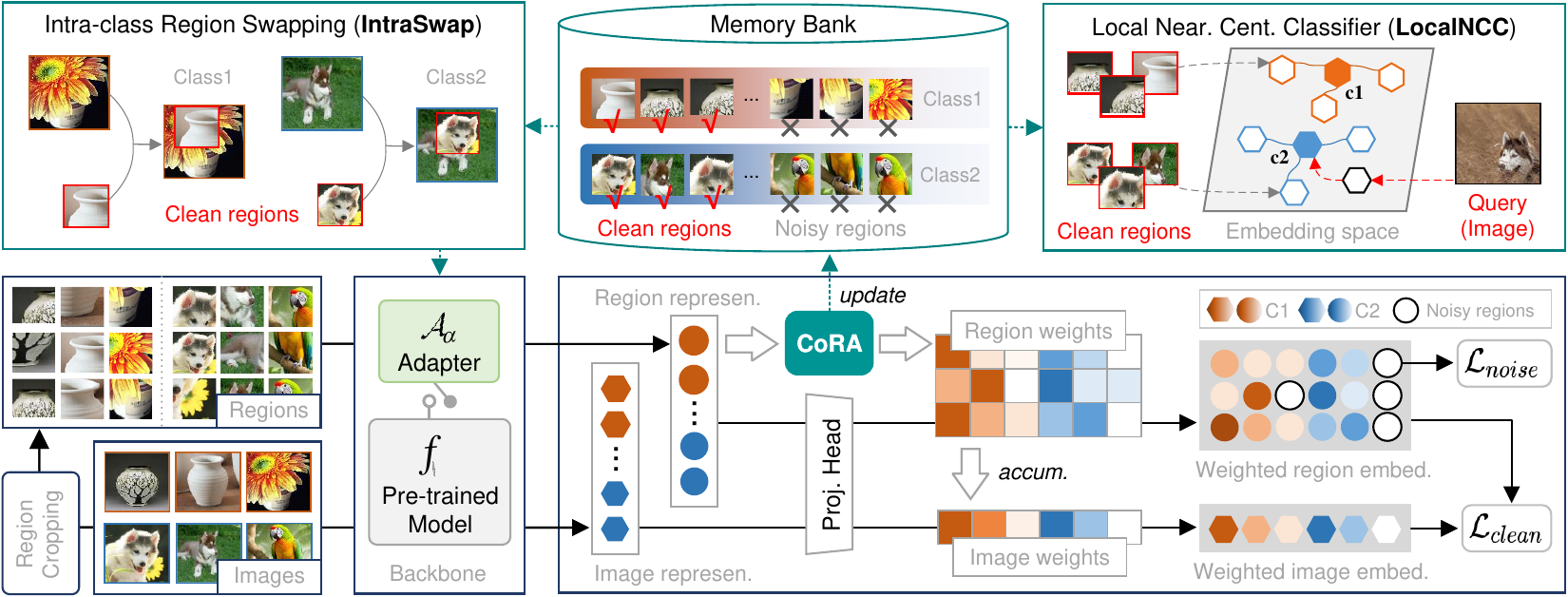} 
	\caption{Overview of our \textbf{DETA}++ framework. 
        {Firstly}, the images together with a set of randomly cropped local regions of support samples are fed into a pre-trained model $f_{{{\theta}}}$ (w/ or w/o a model-specific adapter $\mathcal{A}_{{{\alpha}}}$) to extract image and region representations.
		{Secondly}, a {contrastive relevance aggregation} ({CoRA}) module takes the region representations as input to determine the weight of each region, based on which we can compute the image weights with a momentum accumulator.
	\textcolor{black}{{Thirdly}, a clean prototype loss $\boldsymbol{\mathcal{L}_{clean}}$ and a noise entropy maximization loss $\boldsymbol{\mathcal{L}_{noise}}$ are devised in a low-dimensional embedding space to establish noise-robust task adaptation. }
        {Finally}, we employ a memory bank to store and refine clean regions for each class, based on which an Intra-class Region Swapping ({IntraSwap}) strategy is devised for  prototype rectification during optimization, and a Local Nearest Centroid Classifier ({LocalNCC}) is devised to yield noise-robust predictions on query samples at inference.
        } \label{pip}
\end{figure*}

Despite the encouraging advantages, the vast majority of task adaptation-based FSL approaches have been developed and evaluated in the closed world setting, where both the support and query samples of few-shot tasks are assumed to be clean enough. 
This assumption, however, rarely holds for models deployed in the open world.
As illustrated in Fig. \ref{f1}, both the support and query samples of few-shot tasks collected in the open world can be polluted by in-distribution noise (\textbf{ID noise}) and out-of-distribution noise (\textbf{OOD noise}).
\textcolor{black}{Specifically, ID noise are support/query samples with cluttered image backgrounds. In contrast, OOD noise indicate samples from unknown classes.}
It has been well-recognized that a tiny portion of noisy training samples can compromise the model performance to a large extent \cite{liu2015classification,song2022learning,kong2021resolving}.
When it comes to FSL, 
with limited support samples available, the adverse effect of the dual noises in support samples can be severely amplified during task adaptation. 
\textcolor{black}{
As proven in Fig. \ref{f2}, as the ratio of ID/OOD noise in support samples increases, the generalization ability of the adapted model on target few-shot tasks decreases dramatically.}
\textcolor{black}{Moreover, in the presence of the dual noises, the adapted model may suffer from the overconfidence issue, resulting in unreliable predictions on query samples at inference. This poses a significant threat to the security of AI systems.}
Hence, improving the robustness of the adapted model to the dual noises in few-shot tasks is indispensable for achieving reliable FSL in the open world.
A critical yet underexplored question thus arises:
\begin{framed}
\textcolor{black}{
 Can we address both ID and OOD noise from support and query samples within a unified framework to achieve reliable few-shot learning?}
\end{framed}

\textcolor{black}{
In this work, we answer the above question by proposing {{\textbf{DE}noised  \textbf{T}ask \textbf{A}daptation (\textbf{DETA}++)}}, a first, unified ID- and OOD-denoising framework for reliable FSL. }
The goal of DETA++ is to filter out task-irrelevant region features (e.g. backgrounds) of ID noise and image features of OOD noise, relying only on the local region details of the given support samples.
To this end, DETA++ first leverages a parameter-free Contrastive Relevance Aggregation (CoRA) module to compute the weights of image regions and images by aggregating the task relevance scores of the local regions from support samples. 
\textcolor{black}{Then, based on the calculated region/image weights, two complementary losses, including a \textit{clean prototype} loss ($\mathcal{L}_{clean}$) and a \textit{noise entropy maximization} loss ($\mathcal{L}_{noise}$), are proposed in a low-dimensional embedding space for establishing noise-robust task adaptation. }
Furthermore, DETA++ employs a Memory Bank to store and refine clean regions for each inner-task class at each iteration, based on which an Intra-class Region Swapping (IntraSwap) strategy and a Local Nearest Centroid Classifier (LocalNCC) are devised to improve the robustness of the adapted model to noisy query samples. 
An overview of the DETA++ is presented in Fig. \ref{pip}.
\textcolor{black}{Notably, DETA++ is orthogonal to existing task adaptation-based few-shot classification and OOD detection approaches, therefore can be used as a plugin to improve their robustness against the dual noises. 
We conduct extensive experiments to demonstrate the effectiveness and flexibility of DETA++.
1) \textcolor{black}{We follow the experimental setup of DETA \cite{Zhang_2023_ICCV} to evaluate the ID- and OOD-denoising performance of DETA++ on support samples. }
By performing ID-denoising on the FSL benchmark Meta-Dataset \cite{triantafillou2019meta}, DETA++ improves the ACCs of six strong baselines by \textbf{\underline{2.1}}\%$\sim$\textbf{\underline{5.2}}\% (Table \ref{table1}); by conducting OOD-denoising on the OOD-polluted Meta-Dataset, DETA++ enhances the six baselines by \textbf{\underline{2.7}}\%$\sim$\textbf{\underline{6.6}}\% (Table \ref{table2}). 
2) \textcolor{black}{We assess the ID- and OOD-denoising performance of DETA++ on \textit{query} samples by comparing the OOD detection results of models learned by DETA++ against those trained with other approaches\footnote{\textcolor{black}{The OOD detection performance of a pretrained model not only serves as a direct indicator of its ability to distinguish ID images from OOD noise but also reflects whether the model has learned discriminative representations of ID classes that are robust to ID noise in query samples during inference.}}. }
After applying DETA++ to CoOp$_{\mathrm{MCM}}$ \cite{miyai2024locoop}—a strong few-shot OOD detection baseline, DETA++ reduces the FPR95 by \textbf{\underline{8.6}}\% on the OOD detection benchmark ImageNet-1K, outperforming the previous state-of-the-art by a significant margin (Table \ref{tableood}).}

\textbf{The main contributions of this work are threefold:}
\begin{itemize}
\item We reveal that the overlooked ID and OOD noise in few-shot tasks negatively affect the task adaptation performance of existing FSL methods.
\item \textcolor{black}{We propose DETA++, a first, unified, ID- and OOD-denoising framework for reliable FSL.} DETA++ can be flexibly plugged into existing task adaptation based few-shot classification or OOD detection methods to improve their robustness to the dual noises.
\item We conduct extensive experiments to show the effectiveness and flexibility of DETA++. DETA++ improves competitive baselines and establishes the state-of-the-art performance in both few-shot classification and OOD detection benchmarks.
\end{itemize}

\section{Related Work}
\label{sec:related work}
\keypoint{Few-shot Learning.} 
Few-shot learning (FSL) has made great progress in closing the gap between machine and human intelligence, recently \cite{li2023libfewshot,xiao2022few,wang2021trust,chen2020knowledge}. Prevalent FSL approaches learn new concepts under scarce supervision by a meta-learning setting. Particularly, {optimization-based} schemes \cite{finn2017model,nichol2018first,munkhdalai2017meta} seek to learn a sensitive initialization that can quickly adapt the classifier to the target task with only a few updates. As a well-known optimization-based method, MAML \cite{finn2017model} learns an model initialization that allows fast fine-tuning on few samples. 
Reptile \cite{nichol2018first} and MetaNet\cite{munkhdalai2017meta} are capable of overcoming the instability and slow-generalization problems of MAML.
By contrast, the key idea of {metric learning-based} methods is to learn a transferable embedding space for classifying query samples based on their similarity to each class’s support examples \cite{snell2017prototypical,hou2019cross}. 
The effective ProtoNet \cite{snell2017prototypical} and its variants \cite{ye2020few,chen2019closer} construct the prototypes of classes, and minimizes the distance between queries and the corresponding prototypes. 
Many recent works focus on the learning of task-specific and discriminative feature embeddings. 
FEAT \cite{ye2020few} integrates a set-to-set function ({i.e.}, a Transformer) into ProtoNet to yield task-specific and discriminative feature embeddings. 
CAN \cite{hou2019cross} develops an attention module to highlight the correct region of interest to achieve better matching between support and query samples. 
DeepEMD \cite{zhang2022deepemd} employs the Earth Mover’s Distance as a metric to calculate a structural distance between dense image representations for modeling image relevance. 
As FSL approaches grow in popularity, a growing number of researchers show great interest in the challenging \textit{cross-domain} FSL (CD-FSL) \cite{guo2020broader,tseng2020cross,liang2021boosting,fu2021meta,fu2021metapp}.
Our prior work \cite{zhang2022free} reveals the “style-shift” issue in CD-FSL and performs style-aware episodic training and robust contrastive learning to overcome the issue.
StyleAdv \cite{fu2023styleadv} proposes a novel style attacking method to augment styles leading to a more domain-generalizable FSL model. Compared to the widely-studied (\textit{in-domain}) FSL, CD-FSL is a more realistic yet difficult setting where the available training data from test tasks is not only extremely limited but also presents severe domain difference from  base classes.

\keypoint{Test-time Task Adaptation in FSL.} 
Previous studies in test-time task adaptation \cite{requeima2019fast,bateni2020improved} have revealed that learning task-specific knowledge from test tasks can remarkably boost the generalization of the pre-trained models, especially when there exists severe {category shift} or {domain shift}  between base classes and few-shot tasks. 
The task adaptation approaches CNAPS \cite{requeima2019fast}, Simple CNAPS \cite{bateni2020improved} and FLUTE \cite{triantafillou2021learning} develop task-specific adapters (i.e. FiLM modules) to adapt the pre-trained models to few-shot tasks, relying on the inner-task support samples.
In contrast, \cite{lifchitz2019dense} encourage the 
pre-trained model to learn more discriminative representations by conduct dense classification on few-shot tasks.
The recently proposed URL \cite{li2021universal} employs a pre-classifier feature mapping layer to adapt the feature from a single task-agnostic model learned from multiple domains for target tasks with novel classes. To improve the capacity of URL for task adaptation, 
TSA \cite{li2022cross} adapts the learned feature extractor by building ingenious adapters at multiple ResNet layers. 
\cite{xuexploring} propose a ViT-specific adapter, named eTT, to provide the model with task-specific information.
In recent years, large vision-language pre-trained models (VLPMs), represented by \cite{radford2021learning}, have developed rapidly and achieved great success in many fields.
As a test-time task adaptation technology, \textit{prompt tuning} adapt VLPMs to downstream tasks by learning task-specific prompts  with a handful of labeled training/support examples. Representative CLIP-based prompt tuning are CoOp \cite{zhou2022coop}, CoCoOp \cite{zhou2022conditionalsd}, KgCoOp \cite{yao2023visual}, and etc.
In particular, our recently proposed DePT framework achieved the state-of-the-art performance on both target/base and new tasks by decoupling base-specific and task-shared features during prompt tuning \cite{zhangji2023dept}.

\keypoint{Data-denoising for Robust FSL.}
The training data collected from the open world are unavoidably polluted by image noise or label noise, which may compromise the performance of the learned models \cite{arpit2017closer,song2022learning}.
Limited works in FSL considered the influence of image noise ~\cite{luo2021rectifying} or label noise ~\cite{liang2022few} on model generalization. 
Additionally, they mainly focus on dealing with noises in base classes rather than in the few-shot task.
Particularly, ~\cite{liang2022few} for the first time explored the label noise problem in FSL. Differences between the work \cite{liang2022few} and ours are threefold. 
{i)} We aim to address both ID and OOD noise from the task, where every sample is of great value in characterizing the few-shot task.
{ii)} We take advantage of both global visual information and local region details to achieve the goal.
{iii)} Our method is orthogonal to existing task adaptation based few-shot classification and OOD detection methods.
Even so, ~\cite{liang2022few} does bring a lot of inspiration to our method.
Recently, a plethora of cross-image alignment based FSL methods have been developed to extract more noise-robust representations \cite{zhang2020deepemd,hou2019cross}.
Those methods highlight important local regions by aligning the local features between the support and query samples of few-shot tasks during the meta-training stage. Despite the impressive performance, those {none-adaptation} methods are unable to capture task-specific representations when there exists severe {category shift} or {domain shift}  between base classes and few-shot tasks \cite{luo2023closer,hu2022pushing}.
Moreover, we often overlook the fact that owing to the small sample size in few-shot tasks, negligible computational cost is required to model the relationships of the support samples.


\keypoint{Out-of-distribution Detection.}  
Machine learning systems deployed in the open world can be challenged by out-of-distribution (OOD) data \cite{scheirer2012toward,bendale2015towards,bendale2016towards}. OOD detection seeks to recognize unknown inputs from the open world to prevent unpredictable risks. The vast majority of previous works are {test-time} approaches that rely on the output softmax confidence score of a pretrained model to safeguard against OOD inputs. The insight beneath this line of works is that incoming examples with lower output softmax confidence scores are more likely from OOD \cite{liang2017enhancing,liu2020energy}.
Effective test-time scoring functions include OpenMax \cite{bendale2015towards}, MSP \cite{hendrycks2016baseline}, LogitNorm \cite{wei2022mitigating}, DICE \cite{sun2022dice}, Energy \cite{liu2020energy}, ODIN \cite{liang2017enhancing} and etc.
In the recent work \cite{ahn2023line}, a simple yet effective test-time approach named LINE is proposed. By leveraging important neurons for post-hoc OOD detection, LINE yields remarkable test-time OOD detection performance. Most of those OOD methods consider the development of effective OOD decision functions alone. Our recently proposed MODE  framework considers training-time representation learning and test-time OOD detection, simultaneously \cite{zhangji2023global}.
Recently, a rich line of works are proposed to cope with the challenging few-shot OOD detection problem. The goal of few-shot OOD detection is to detect OOD samples that the model has not been exposed to during training using only a few labeled support samples of the target task  \cite{miyai2024locoop,NEURIPS2021_c91591a8}. 

\keypoint{\textcolor{black}{Differences between DETA \cite{Zhang_2023_ICCV} and DETA++.}}\textcolor{black}{
The proposed DETA++ in this work is an extension of our previous work {DETA} (ICCV'23) \cite{Zhang_2023_ICCV}.  
{The new contributions in this work are significant for the following reasons:}
\begin{itemize}
\item  \textcolor{black}{\textbf{Idea}: DETA tackles image and label noise in support samples, while DETA++ addresses \textcolor{black} ID and OOD noise in {both} support and query samples.}
\item  \textbf{Methodology}: 
To address both ID and OOD noise in a unified framework, DETA++ differs from DETA in the following aspects. 
\textcolor{black}{1) During training, DETA devises a local compactness loss $\mathcal{L}_l$ and a global  dispersion loss $\mathcal{L}_g$ based on weighted regions/images to improve model robustness to image and label noise. In contrast, DETA++ proposes a clean prototype loss $\mathcal{L}_{clean}$ and a noise entropy maximization loss $\mathcal{L}_{noise}$ to jointly overcome the adverse impacts of ID and OOD noise. 
More specifically, $\mathcal{L}_{clean}$ is an optimized version of $\mathcal{L}_g$, designed to pull clean regions closer to the their class prototypes constructed from clean images, while $\mathcal{L}_{noise}$ is proposed to enhance the decision boundary between class features and noise features by maximizing the entropy of model predictions on noisy regions recognized by the CoRA module. 
Quantitative comparisons of $\mathcal{L}_l$, $\mathcal{L}_g$, $\mathcal{L}_{clean}$ and $\mathcal{L}_{noise}$ are presented in Section \ref{sect.losscompare}.}
2) At inference, DETA designs a Nearest Centroid Classifier (NCC) using weighted, high-dimensional image features, while DETA++ designs a Local NCC (LocalNCC) using clean, low-dimensional region embeddings stored in a memory bank. 
3) Based on the stored clean regions in a memory bank, DETA++ further designs an Intra-class Region Swapping (IntraSwap) strategy to rectify ID class prototypes, thereby improving the model’s robustness to the dual noises. 
\item  \textbf{Experiments}: 
\textcolor{black}{On the one hand, we follow the experimental setup of DETA to evaluate the effectiveness of DETA++ in tackling the dual noises in \textit{support} samples.}
From the obtained results on the Meta-Dataset benchmark, DETA++ consistently improves the performance and outperforms DETA on the six baselines. 
\textcolor{black}{On the other hand, we assess the effectiveness of DETA++ in tackling the dual noises in \textit{query} samples by comparing the OOD detection results of models learned by DETA++ against those trained with other approaches.} 
From the achieved results on the challenging ImageNet-1K benchmark, DETA++ outperforms DETA and a wide spectrum of state-of-the-arts by significant margins.
\end{itemize}}

\section{Proposed Approach}
In this section, we elaborate on our proposed DEnoised Task Adaptation (DETA++) approach. 

\subsection{Preliminaries}
\label{pre}
Assume we have a pre-trained model $f_\theta$ parameterized by ${{\theta}}$, which serves as a feature backbone to output a \textit{d}-dimensional representation for each input image.
Task adaptation based FSL seeks to adapt $f_{\theta}$ to the target task $T=\{S,Q\}$, by mining task-specific knowledge from the few-labeled support samples ${S}=\{(\boldsymbol{x}_i,y_i)\}_{i=1}^{N_s}$, consisting of $N_s$ {{image}}-{{{label}}} pairs from $C$ novel classes, {i.e.}, $y_i \in \{1,..., C\}$.
It is expected that the adapted model can correctly partition the $N_q$ query samples ${{Q}}=\{(\boldsymbol{x}_i)\}_{i=1}^{N_q}$ to the $C$ classes in the {representation} space. 
If there are exactly $K$ support samples in each of these $C$ classes, the task is also called a $C$-way $K$-shot task.

\keypoint{Adapter-finetuning based Task Adaptation}. 
The goal of adapter-finetuning (adapter-FT) based task adaptation is to capture the knowledge of the target task by attaching a model-specific adapter $\mathcal{A}_{{{\alpha}}}$ parameterized by ${{\alpha}}$ to the pre-trained model $f_{{{\theta}}}$. 
During task adaptation, the parameters of $f_{{{\theta}}}$, i.e. ${\theta}$, are frozen and only the adapter parameters ${\alpha}$ are optimized from scratch on the given support samples:
\begin{equation} 
\alpha := \alpha - \gamma \nabla_\alpha \mathcal{L}^{{S}}\big([{\theta};{\alpha}]\big),
\end{equation}
where $\gamma$ is the learning rate, and
\begin{equation} 
 \mathcal{L}^{{S}}\big([{\theta};{\alpha}]\big)=\frac{1}{N_s}\sum_{(\boldsymbol{x},y) \in {S}}\ell\big(h(f_{[{\theta};{\alpha}]}(\boldsymbol{x});S), y\big),
\end{equation}
where $\ell$ indicates cross-entropy loss, $f_{[{\theta};{\alpha}]}$ is the adapter-integrated feature backbone, and $h$ is a non-parametric classifier head capable of producing a softmax probability vector whose dimensionality equals $C$.
Notably, the adapter-FT based method TSA \cite{li2022cross} achieved remarkable results on the Meta-Dataset \cite{triantafillou2019meta} benchmark, by integrating a residual-adapter to the pre-trained URL \cite{li2021universal} model, and setting $h$ to the nonparametric Nearest Centroid Classifier (NCC) \cite{mensink2013distance}\footnote{The non-parametric NCC has been proven in \cite{li2022cross} to be more effective for adapter-finetuning or full-finetuning based task adaptation than other competitors such as logistic regression, SVM, e.t.c.}.

\keypoint{Full-finetuning based Task Adaptation}. Adapter-FT based task adaptation approaches need to develop model-specific adapters to adapt different pre-trained models, {e.g.,} the residual adapter TSA \cite{li2022cross} for ResNets \cite{he2022masked}, the self-attention adapter eTT \cite{xuexploring} for ViTs \cite{dosovitskiy2020image}. 
In contrast, full-finetuning (full-FT) based task adaptation, originated from transfer learning literature ~\cite{kornblith2019better}, finetunes all pre-trained parameters of $f_{{{\theta}}}$ via $\theta := \theta - \gamma \nabla \mathcal{L}^{{S}}(\theta)$. 
Thus, full-FT based methods are model-agnostic, and can be more flexible compared to adapter-based methods.

\subsection{Overview}
\textcolor{black}{The goal of our DETA++ is to overcome the adverse effects of both ID and OOD noise from support and query samples within a unified framework, thereby achieving reliable few-shot learning.}
To this end, DETA++ filters out task-irrelevant region features (e.g., image backgrounds) of ID noise together with image features of OOD noise, by taking advantages of the local region details of the available support samples. 
An overview of DETA++ is shown in Fig. \ref{pip}. 
Concretely, {for each iteration of the task adaptation process:
\begin{itemize}
\item   A pre-trained model $f_{{{\theta}}}$ (w/ or w/o a model-specific adapter $\mathcal{A}_{{{\alpha}}}$) takes the images and a set of randomly cropped image regions of support samples as input to obtain image and region features. 

\item   A parameter-free {Contrastive Relevance Aggregation} (CoRA) module takes the region features as input to calculate the weights of image regions, based on which we can compute the weights of images with a momentum accumulator.

\item   \textcolor{black}{A projection head maps the high-dimensional image and region representations to a low-dimensional embedding space, where a clean prototype loss ${\mathcal{L}_{clean}}$ and a noise entropy maximization loss ${\mathcal{L}_{noise}}$ are devised to achieve noise-robust task adaptation.} 

\item  A memory bank is used to store and refine clean regions for each ID class, based on which an Intra-class Region Swapping ({IntraSwap}) strategy is developed for prototype rectification during optimization.

\end{itemize}

At inference, a noise-robust Local Nearest Centroid Classifier (LocalNCC) is devised based on the refined clean regions of each class in the memory bank.

\subsection{Contrastive Relevance Aggregation (CoRA)}
\label{cora}
The motivation of the CoRA module is that image regions, which show higher relevance to \textit{in-class} regions while lower relevance to \textit{out-of-class} regions, are more likely to be object regions, therefore should be assigned larger weights. On the contrary, image regions, which show lower relevance to \textit{in-class} regions while higher relevance to \textit{out-of-class} regions, are more likely to be noisy regions from ID or OOD noise, thus should be assigned lower weights. 

Given the support samples ${S}=\{(\boldsymbol{x}_i,y_i)\}_{i=1}^{N_s}$ of a test-time task, we first randomly crop $k$ $\Omega$$\times$$\Omega$-size local regions for every image $\boldsymbol{x}_i$.
Next, the original image together with all cropped regions of each sample are fed into $f$ to generate the image representation $\boldsymbol{{{z}}}_i$ and region representations $Z_i =\{\boldsymbol{{{z}}}_{ij}\}_{j=1}^{k}$. 
Let $Z^{(c)}=\bigcup_{i=1}^{N_{c}}Z_i$ denote the collection of representations of cropped regions in class $c$, $\textbf{\textit{Z}}=\bigcup_{c=1}^{C}Z^{(c)}$ the set of representations of cropped regions, where $N_c$ is the number of images in class $c$. 
For each region representation $\boldsymbol{{{z}}}_{ij}$ in $Z_i$, we construct its \textit{in-class} and \textit{out-of-class} region representation sets as 
${I}(\boldsymbol{z}_{ij})=Z^{(c)}\setminus Z_i$ and ${O}(\boldsymbol{z}_{ij})=\textbf{\textit{Z}}\setminus Z^{(c)}$, respectively.
Note that in $I(\boldsymbol{z}_{ij})$, the other $k-1$ intra-image representations are dropped to alleviate their dominating impacts. 
CoRA calculates the weight of each region based on the global statistics of in-class and out-of-class relevance scores, respectively formulated as
\begin{equation}
        \phi(\boldsymbol{z}_{ij}) = \frac{1}{|I(\boldsymbol{z}_{ij})|} \sum_{\boldsymbol{{{z}}}' \in I(\boldsymbol{z}_{ij})} \zeta(\boldsymbol{{{z}}}_{ij}, \boldsymbol{{{z}}}'),
        \label{eq1}
\end{equation}
\begin{equation}
        \psi(\boldsymbol{z}_{ij})
         = \frac{1}{|O(\boldsymbol{z}_{ij})|} \sum_{\boldsymbol{{{z}}}' \in O(\boldsymbol{z}_{ij})} \zeta(\boldsymbol{{{z}}}_{ij}, \boldsymbol{{{z}}}'),
         \label{eq2}
\end{equation}
\noindent where $\zeta(\cdot)$ indicates cosine similarity.
\textcolor{black}{In order to identify the most discriminative representations for every ID class, those scores are then normalized inside the class:}
\begin{equation}
    \widetilde{\phi}(\boldsymbol{z}_{ij})\!=\!\frac{e^{{\phi(\boldsymbol{z}_{ij})}}}{\sum_{\substack{\boldsymbol{{{z}}}' \in Z^{(c)}}}{\!e^{\phi(\boldsymbol{{{z}}}')}}},
\end{equation}
\begin{equation}
\widetilde{\psi}(\boldsymbol{z}_{ij})\!=\!\frac{e^{{\psi(\boldsymbol{z}_{ij})}}}{\sum_{\substack{\boldsymbol{{{z}}}' \in Z^{(c)}}}{\!e^{\psi(\boldsymbol{{{z}}}')}}}.
\end{equation}
Therefore, the region weight for $\boldsymbol{z}_{ij}$ can be defined as 
\begin{equation}
\lambda_{ij}=\widetilde{\phi}(\boldsymbol{z}_{ij})/\widetilde{\psi}(\boldsymbol{z}_{ij})\in\mathbb{R}. 
\end{equation}

Finally, we can easily use a threshold $\varrho$ to separate clean and noisy regions during each iteration. Denote $\textbf{\textit{R}}^t$ as a set of cropped image regions in the $t$-th iteration, the recognized noisy regions can be expressed as $\mathcal{N}^t=\{o_i \in \textbf{\textit{R}}^t : \lambda_{i} < \varrho\}$, where ${\lambda}_{i}$ is the weight of the \textit{i}-th region computed by CoRA. Note that $\mathcal{N}^t$ may consist of task-irrelevant image regions from both ID and OOD noise.
Therefore, the set of recognized clean image regions in the $t$-th iteration can be expressed as $\mathcal{C}^t = \textbf{\textit{R}}^t \setminus \mathcal{N}^t$.

\keypoint{A Momentum Accumulator for Weighting Images.}
Aside from weighting the local regions of images from support samples, we also need to assess the {quality} of the images themselves for filtering out OOD noise in support samples.

Intuitively, the most direct way to determine the weight of an image $\boldsymbol{x}_i$, denoted as $\omega_i$, is to average the weights of all $k$ cropped regions belonging to it, {i.e.}, 
\begin{equation}
\omega_i = \frac{1}{k} \sum_{j=1}^k \lambda_{ij}.
\end{equation}
However, the randomly cropped regions at different iterations may have large variations, resulting the frailty of the calculated image weights.
A Momentum Accumulator (MA) is thus developed to cope with this limitation, which takes the form of
\begin{equation}
    \omega^{t}_i =
\begin{cases} 
\frac{1}{k} \sum_{j=1}^k  \lambda_{ij}, & \mbox{if }t=1 \\
\gamma \omega^{t-1}_i+ \frac{1-\gamma}{k} \sum_{j=1}^k   \lambda_{ij}, & \mbox{if } t>1
\end{cases}
\end{equation}
where $\omega^{t}_i$ denotes the accumulated image weight of $\boldsymbol{x}_i$ in the $t$-th iteration of task adaptation, and $\gamma$ is the momentum hyperparameter, and we set it to 0.7 in our method.
We omit the superscript $t$ in the following sections for brevity.


\subsection{Noise-robust Task Adaptation}

To achieve noise-robust task adaptation, we design a learning objective that enables an discriminative embedding space, which satisfies two properties: \textbf{\textcircled{1}} \textbf{each ID support/query sample has a higher probability assigned to the correct ID class}, and \textbf{\textcircled{2}} \textbf{each OOD support/query sample has a lower probability assigned to any ID class}.
\textcolor{black}{Following the common practice for contrastive representation learning \cite{chen2020simple,chen2022learning,rusu2018meta},} we first project each image feature $\boldsymbol{{{z}}}_i$ and its region features ${Z}_i=\{\boldsymbol{{{z}}}_{ij}\}_{j=1}^{k}$ into a low-dimensional embedding space using a projection head, and denote the $l_2$-normalized image embedding and $k$ region embeddings as $\boldsymbol{{{e}}}_i$ and  ${E}_i=\{\boldsymbol{e}_{ij}\}_{j=1}^{k}=\{\boldsymbol{r}_{\iota}\}_{\iota=1}^{k}$, respectively. 

\textcolor{black}{\underline{\textbf{To achieve \textbf{\textcircled{1}}}}, 
we propose a {clean prototype} loss that promotes {\textit{clean}} regions to be close to their image-level class prototypes guided by the calculated image/region weights of the CoRA module.}
Specifically, we first use all clean image embeddings in $\{\boldsymbol{e}_{i}\}_{{}=1}^{N_s}$ to construct $C$ weighted class prototypes as follows:
\begin{equation} 
	\begin{aligned}
	 &\boldsymbol{\mu}_c = \frac{1}{N_c} \sum_{y_{i}=c} \omega_{i} \boldsymbol{e}_{i}, \,\,\, c = 1,2, ..., C, \\
     &\mathrm{s.t.} \,\, \omega_{i} =0 \, \, \mathrm{if} \, \,  \omega_{i}<\varrho,
	\end{aligned}
	\label{eq.proto}
\end{equation}
where $\omega_{i}$ is the weight of $i$-th image\footnote{We use the same threshold $\varrho$ to filter out noisy regions and noisy images for brevity.}.
Next, we can estimate the likelihood of every region embedding $\boldsymbol{r}_{j}$, based on a softmax over distances to the class prototypes:
\begin{equation}
		p(y=m|\boldsymbol{r}_{j}) = \frac{\exp\big(\zeta(\boldsymbol{r}_{j},\boldsymbol{\mu}_{m})\big)}{\sum_{c=1}^C\exp\big(\zeta(\boldsymbol{r}_{j},\boldsymbol{\mu}_c)\big)},
  \label{e10}
\end{equation}
where $\zeta(\cdot)$ is cosine similarity. 
Thus, the clean prototype loss, denoted as $\mathcal{L}_{clean}$, can be expressed as 
\begin{equation}
\begin{aligned}
      {  {\mathcal{L}_{{clean}}} } &=  - \frac{1}{N_s\times k} \sum_{{i}=1}^{N_s\times k} \lambda_{i} \log\big(p(y=y_{i}|\boldsymbol{r}_{i})\big), \\
        &\mathrm{s.t.} \,\, \lambda_{i} =0 \, \, \mathrm{if} \, \,  \lambda_{i}<\varrho,
        \label{ee}
\end{aligned}
\end{equation}
where $\lambda_{i}$ is used to constrain the impact of the $i$-th image region.
In this way, the adverse effect of task-irrelevant local features of ID noise and global features of OOD noise from support samples can be mitigated during optimization\footnote{\textcolor{black}{$\mathcal{L}_{clean}$ is an optimized version of the global dispersion loss $\mathcal{L}_g$ devised in DETA \cite{Zhang_2023_ICCV}, which pulls all weighted regions closer to their corresponding class prototypes without rigorously  filtering out noisy regions using the threshold $\varrho$.}}.

\textcolor{black}{\underline{\textbf{To achieve \textbf{\textcircled{2}}}}, we propose a noise entropy maximization loss (i.e., $\mathcal{L}_{noise}$) that enhances the decision boundary between class features and noise features by maximizing the entropy of model predictions on noisy regions recognized by the devised CoRA module. }
The motivation is that as the distribution space for potential OOD noise in the open world can be prohibitively large, it is critical to leverage the detected noisy regions to improve the model's ability in recognizing OOD noise/samples during the training process.

\textcolor{black}{Specifically, the devised $\mathcal{L}_{noise}$ maximizes the entropy of model predictions on the detected noisy regions (in $\mathcal{N}^t$) to push the embeddings of ID/OOD noise far away from the embeddings of inner-task classes. } Formally, 
\begin{equation}
\begin{aligned}
       {  \mathcal{L}_{noise} = - \frac{1}{|\mathcal{N}^t|} \sum_{i=1}^{|\mathcal{N}^t|} \mathcal{E}(p_i), }
\end{aligned}
\end{equation}
where {$\mathcal{E}(\cdot)$} is the entropy function, $p_i$ indicates the probabilities of the $i$-th noisy region belonging to the $C$ ID classes, determined by Eq. (\ref{e10}).



Finally, the devised two losses complement each other to tackles the dual noises in a unified framework, by optimizing the following objective:
	\begin{equation} 
	\begin{aligned}
	 \mathcal{L} &= \mathcal{L}_{clean} +  \beta \mathcal{L}_{noise}, \\
	\end{aligned}
	\label{eq12}
	\end{equation}
where the coefficient $\beta$ is used to control the contribution of $\mathcal{L}_{noise}$.
During task adaptation, we iteratively construct a set of local regions from the support samples of the target task, and perform SGD update using $\mathcal{L}$. 
\textcolor{black}{
The training pipeline of the proposed DETA++ framework is shown in  \textbf{Algorithm 1}. Remarkably, DETA++ can be used as a plugin to improve both adapter-finetuning based and full-finetuning based task adaptation methods. }

\begin{algorithm}[t]
	\caption{The training pipeline of DETA++}
	\label{alg:algorithm1}
	\KwIn{The target task $T=\{S,Q\}$; A pretrained model $f_{\theta}$;
	A model-specific adapter $\mathcal{A}_{{{\alpha}}}$ for adapter-FT based task adaptation; 
    MaxIteration $\eta$; Learning rate $\gamma$.}  
    \BlankLine
   	\# full-FT based Task Adaptation (\textbf{Case 1})\\
	\While{\textnormal{not reach $\eta$}}{
		Crop a collection of image regions $R$ from $S$; \\
		Compute region weights by CoRA; \\  
	    Refine image weights in the accumulator; \\
        Refine clean regions in the memory bank; \\
		Update $\theta$ by $\mathcal{L}$. 
	}
    \KwOut{Optimized $\theta^*$.}
	\BlankLine
	\# adapter-FT based Task Adaptation (\textbf{Case 2}) \\
	Froze $\Theta$, and initialize ${\alpha}$;\\
	\While{\textnormal{not reach $\eta$}}{
		Crop a collection of image regions $R$ from $S$; \\
		Calculate region weights by CoRA; \\  
	    Refine image weights in the accumulator; \\
        Refine clean regions in the memory bank; \\
		Update ${\alpha}$ by $\mathcal{L}$. 
	}
    \KwOut{Optimized ${\alpha^*}$.}
\end{algorithm}

\subsection{Further Exploring the “Free-lunch” in CoRA}
Apart from establishing noise-robust task adaptation with Eq. (\ref{eq12}), the devised CoRA module can be further explored to boost the robustness of DETA++ under the dual noises.

\keypoint{A Memory Bank for Storing and Refining Clean Regions.}
According to Section \ref{cora}, the set of clean regions for the task can be defined as $\boldsymbol{\mathcal{C}}=\bigcup_{t=1}^{\eta}\mathcal{C}^t$, where $\eta$ is the number of iterations.
During each iteration, a Memory Bank is used to store the image regions with the top-2\textit{K} weights for each class, where $K$ is the number of training samples per class. 
For computational efficiency, all image regions are stored as bounding box coordinates.

\keypoint{IntraSwap: Intra-class Region Swapping for Prototype Rectification.}
Training on narrow-size distribution of scarce data usually tends to get biased class prototypes \cite{liu2020prototype,xu2022alleviating}. 
This challenge becomes greater when the support samples are polluted by ID and OOD noise.
As a result, it greatly limits the effectiveness and robustness of the proposed DETA++.
One direct way to cope with this problem is to rectify the class prototypes of ID classes in Eq. (\ref{eq.proto}), by augmenting the distribution of the support set with clean images during task adaptation.

As a representative distribution augmentation strategy, CutMix \cite{yun2019cutmix} has shown to be capable of preventing overfitting and improving the generalization of the learned model. 
Let $\boldsymbol{x}\in \mathbb{R}^{W\times H\times C}$ and $y$ denote a training image and its label, respectively. CutMix generates a new training image ($\boldsymbol{x}^*$, $y^*$) by cutting and pasting a random image region from one image ($\boldsymbol{x}_{\mathrm{A}}$, $y_{\mathrm{A}}$)  onto another image ($\boldsymbol{x}_{\mathrm{B}}$, $y_{\mathrm{B}}$).
The combining operation is as follows: 
\begin{equation} 
	\begin{aligned}
	 \boldsymbol{x}^* &= \mathrm{\textbf{M}} \odot \boldsymbol{x}_{\mathrm{A}} +  (\textbf{1}- \mathrm{\textbf{M}}) \odot \boldsymbol{x}_{\mathrm{B}}, \\
     {y}^* &= \varpi y_{\mathrm{A}} + ({1}-\varpi) y_{\mathrm{B}},
	\end{aligned}
	\label{eq.cutmix}
\end{equation}
where $\mathrm{\textbf{M}} \in \{0, 1\}^{W\times H}$ indicates where to drop out and fill in from $\boldsymbol{x}_{\mathrm{A}}$ and $\boldsymbol{x}_{\mathrm{B}}$, \textbf{1} is a binary
mask filled with ones, and $\varpi$ is usually sample from the uniform distribution (0, 1).

Despite the remarkable advantages, CutMix could potentially lead to a loss of valuable information in the original images, especially when the object regions are replaced by noisy regions. 
To cope with this, we propose Intra-class Region Swapping (IntraSwap), which performs prototype rectification for ID classes by pasting clean regions (in the memory bank) on images from the same class and preserving the corresponding labels of the images.
According to Eq.(\ref{eq.cutmix}), IntraSwap combines two images from the same class, i.e. ($\boldsymbol{x}_{\mathrm{A}}$, $y$) and ($\boldsymbol{x}_{\mathrm{B}}$, $y$), by
\begin{equation} 
	\begin{aligned}
	 \boldsymbol{x}^* &= \mathrm{\textbf{M}} \odot \boldsymbol{x}_{\mathrm{A}} +  (\textbf{1}- \mathrm{\textbf{M}}) \odot \boldsymbol{x}_{\mathrm{M}}, \quad {y}^* = y, \\
	\end{aligned}
	\label{eq.IntraCutMix}
\end{equation}
where $\mathrm{\textbf{M}} \in \{0, 1\}^{W\times H}$ is determined by the stored bounding box coordinates of clean regions with the same label $y$ in the memory bank.
To avoid the loss of the structural information in each image, during task adaptation, we do not perform region cropping on the generated images as on the original images in the CoRA module. This indicates that the generated images are solely used to produce global features to rectify the class prototypes in Eq. (\ref{eq.proto}).

\begin{algorithm}[t]
	\caption{Few-shot classification w/ LocalNCC}
	\label{alg:algorithm1}
	\KwIn{The labeled support set $S$; clean regions of each class in the memory bank; an unlabelled query image $x_i$; the learned embedding $\psi$; the selected similarity metric $\zeta$.}  
    \BlankLine
   	Compute the centroid of clean regions $\boldsymbol{c}_n$ for all $n=1,...,C$ classes in the embedding space; \\
    Compute $\zeta(\boldsymbol{c}_n, \psi(x_i))$; \\
    Predict via $\boldsymbol{\arg \max}_n \zeta(\boldsymbol{c}_n, \psi(x_i))$. \\
    \KwOut{The predicted label of $x_i$.}
\end{algorithm}



\keypoint{LocalNCC: Local Nearest Centroid Classifier for Noise-robust Inference.}
As mentioned before, the adapted model can produce unreliable predictions on query samples when faced with ID and OOD noise. To tackle this issue, we use the top-2\textit{K} clean regions of each class stored in the memory bank to build a Local Nearest Centroid Classifier (LocalNCC) for achieving noise-robust predictions on query images at inference.
Unlike DETA \cite{Zhang_2023_ICCV} which uses global images to build a NCC in a high-dimensional representation space, DETA++ leverages clean image regions to construct a local NCC in a learned low-dimensional embedding space, which can be more conducive to eliminating the adverse effect of the dual noises in query samples.
An illustration of LocalNCC is shown in \textbf{Algorithm 2} (and Fig. \ref{pip}).

\section{Experiments}
\textcolor{black}{In this section, we perform extensive experiments on both few-shot classification and few-shot OOD detection benchmarks to evaluate the effectiveness and flexibility of our DETA++ in overcoming the adverse effects of ID and OOD noise in FSL tasks. 
In Section \ref{supp}, we follow the experimental setup of DETA \cite{Zhang_2023_ICCV} and employ Meta-Dataset \cite{triantafillou2019meta}, one of the most comprehensive few-shot classification benchmarks, to demonstrate DETA++'s effectiveness in tacking the dual noises in support samples.
During inference, the OOD detection performance of a learned model not only serves as a direct indicator of its ability to distinguish ID images from OOD noise but also reflects whether the model has learned discriminative characteristics of ID classes that are robust to ID noise in query samples. In Section \ref{fsood}, we therefore compare the OOD detection performance of models learned by DETA++ against those trained with other approaches on the challenging few-shot OOD detection benchmark ImageNet-1K \cite{deng2009imagenet} to validate the effectiveness of DETA++ in addressing the dual noises in query samples. 
Qualitative and quantitative comparisons of the losses used in DETA and DETA++ are provided in Section \ref{sect.losscompare}. 
}

\subsection{Evaluation on Few-shot Classification} \label{supp}
\textcolor{black}{In line with DETA \cite{Zhang_2023_ICCV}, we apply DETA++ to six few-shot classification baselines to assess its ID- and OOD-denoising capabilities on support samples.}

\begin{table*}[htbp]
\setlength{\abovecaptionskip}{0.01cm}  
  \centering
  \tabcolsep 0.076in
  \caption{Few-shot classification ACCs (\%) of six baseline methods w/ or w/o DETA \cite{Zhang_2023_ICCV} and DETA++ on the \textbf{vanilla Meta-Dataset} (\textbf{w/o OOD noise}). Both {TSA}\cite{li2022cross} and {{eTT}} \cite{xuexploring} are adapter-finetuning based methods, while other baselines are full-finetuning based methods. The best and second best results on each baseline method are \textbf{bold}  and \underline{underlined}, respectively.}
    \begin{tabular}{l|l|cccccccccc|c}
    \hline
     \multicolumn{1}{l|}{Method} & Pret.Model &ImageNet & Omglot & Acraft&CUB& DTD& QkDraw& Fungi& Flower& COCO&Sign& {\textbf{Avg}} \\
    \hline 
     {TSA} \cite{li2022cross} & URL \cite{liu2020universal}  & 58.3\tiny{±0.9}  &80.7\tiny{±0.2}  &61.1\tiny{±0.7} &83.2\tiny{±0.5}& 72.5\tiny{±0.6} & 78.9\tiny{±0.6} & 64.7\tiny{±0.8} & 92.3\tiny{±0.3}  & 75.1\tiny{±0.7}& 87.7\tiny{±0.4}  & 75.5   \\
     \rowcolor{gray!12} {{DETA}} \cite{Zhang_2023_ICCV}  &  &  \textbf{58.7}\tiny{±0.9}  & \underline{82.7}\tiny{±0.2} & \underline{63.1}\tiny{±0.7} & \textbf{85.0\tiny{±0.5}}&  \underline{72.7}\tiny{±0.6} &  \textbf{80.4}\tiny{±0.6} &  \underline{66.7}\tiny{±0.8} &  \textbf{93.8\tiny{±0.3}}  &  \underline{76.3}\tiny{±0.7}&  \textbf{92.1\tiny{±0.4}}  &  \underline{77.3} (\textcolor{red}{{+1.8}})  \\
     \rowcolor{green!8} \textbf{{DETA}}++  &  & \underline{58.5}\tiny{±0.9}    & \textbf{83.5}\tiny{±0.3}  &  \textbf{64.7}\tiny{±0.7} & \underline{84.7}\tiny{±0.5} & \textbf{73.6}\tiny{±0.5}  & \textbf{80.4}\tiny{±0.6} & \textbf{67.8}\tiny{±0.9} & \underline{93.5}\tiny{±0.3}  & \textbf{79.7}\tiny{±0.7} & \underline{91.9}\tiny{±0.4} & \textbf{77.6} (\textcolor{red}{{+2.1}})  \\
     \hline
      {eTT} \cite{xuexploring} & DINO \cite{caron2021emerging}  & 73.2\tiny{±0.8}  & \underline{93.0}\tiny{±0.4} & \underline{68.1}\tiny{±0.7} & 89.6\tiny{±0.3}& 74.9\tiny{±0.5}  & 79.3\tiny{±0.7} & 76.2\tiny{±0.5}  & \underline{96.0}\tiny{±0.2} & 72.7\tiny{±0.6}  & 86.3\tiny{±0.7}  & 80.9 \\
      \rowcolor{gray!12}  {{DETA}} \cite{Zhang_2023_ICCV} & & \underline{75.6}\tiny{±0.8} & \textbf{93.6\tiny{±0.4}} & {67.7\tiny{±0.8}} & \textbf{91.8\tiny{±0.3}} & \underline{76.0}\tiny{±0.5} & \underline{81.9}\tiny{±0.7} & \underline{77.2}\tiny{±0.5} & \textbf{96.9\tiny{±0.3}} & \underline{78.5}\tiny{±0.6} & \underline{88.5}\tiny{±0.7} &   {\underline{82.8}} (\textcolor{red}{{+1.9}}) \\
     \rowcolor{green!8} \textbf{{DETA}}++ &  &  \textbf{78.3}\tiny{±0.8}    &  92.7\tiny{±0.4} &  \textbf{68.6}\tiny{±0.8} & \underline{91.6}\tiny{±0.3} & \textbf{79.5}\tiny{±0.5}  & \textbf{83.1}\tiny{±0.8} & \textbf{77.6}\tiny{±0.5} & 95.8\tiny{±0.3}  & \textbf{80.5}\tiny{±0.6} & \textbf{88.7}\tiny{±0.7} & \textbf{83.6} (\textcolor{red}{{+2.7}}) \\
    \hline
    {full-FT} \cite{Zhang_2023_ICCV}   & MoCo \cite{he2020momentum} &   70.7\tiny{±1.0} &82.5\tiny{±0.4} & 55.1\tiny{±0.8} & 67.0\tiny{±0.8} &   {\textbf{81.3\tiny{±0.5}}} & 73.8\tiny{±0.7} & 54.8\tiny{±0.9} & 89.2\tiny{±0.5}& 76.8\tiny{±0.7}  & 79.6\tiny{±0.6}  & 73.0  \\
    \rowcolor{gray!12}  {{DETA}} \cite{Zhang_2023_ICCV}  &  &     {\underline{73.6}\tiny{±1.0}}  &   {\underline{83.9}\tiny{±0.4}}  &    {\underline{59.1}\tiny{±0.8}} &    {\textbf{73.9\tiny{±0.8}}}  &  \underline{80.9}\tiny{±0.5} &    {\underline{76.1}\tiny{±0.7}}  &    {\textbf{60.7\tiny{±0.9}}} &    {\underline{92.3}\tiny{±0.5}} &   {\underline{79.0}\tiny{±0.7}}    &    {\underline{84.2}\tiny{±0.6}}   &    {\underline{76.4}} (\textcolor{red}{{+3.4}})  \\
    \rowcolor{green!8} \textbf{{DETA}}++ &   & \textbf{75.2}\tiny{±0.9} &  \textbf{87.2}\tiny{±0.4} & \textbf{60.3}\tiny{±0.8}  & \underline{73.1}\tiny{±0.8} &  80.4\tiny{±0.5} & \textbf{78.8}\tiny{±0.8} & \underline{59.9}\tiny{±0.9} &  \textbf{94.0}\tiny{±0.5} & \textbf{82.4}\tiny{±0.7} & \textbf{86.7}\tiny{±0.6} &  \textbf{77.8} (\textcolor{red}{{+4.8}}) \\
    \hline  
     {full-FT} \cite{Zhang_2023_ICCV}   &  CLIP \cite{radford2021learning} & 67.0\tiny{±1.0} &89.2\tiny{±0.5} &  \underline{61.2}\tiny{±0.8} & 84.0\tiny{±0.7} & 74.5\tiny{±0.6} & 75.5\tiny{±0.7} & 57.6\tiny{±0.9} & 92.1\tiny{±0.4}  & 72.1\tiny{±0.8}& 79.8\tiny{±0.7}  & 75.3 \\
    \rowcolor{gray!12} {{DETA}} \cite{Zhang_2023_ICCV}  & &  {\textbf{69.6\tiny{±0.9}}} &  {\underline{92.2}\tiny{±0.5}} &   {{59.7\tiny{±0.8}}} &   {\textbf{88.5\tiny{±0.7}}} &  {\underline{76.2}\tiny{±0.6}} &  {\underline{77.2}\tiny{±0.7}} &  {\underline{64.5}\tiny{±0.9}}  &   {\textbf{94.5\tiny{±0.3}}} &   {\underline{72.6}\tiny{±0.8}}  &  {\underline{80.7}\tiny{±0.7}} &  {\underline{77.6}} (\textcolor{red}{{+2.3}}) \\ 
    \rowcolor{green!8}  \textbf{{DETA}}++ &  & \underline{68.8}\tiny{±0.9}   & \textbf{92.7}\tiny{±0.5}  & \textbf{61.6}\tiny{±0.8}  & \underline{87.7}\tiny{±0.7} & \textbf{78.5}\tiny{±0.6}  & \textbf{80.4}\tiny{±0.7} & \textbf{68.1}\tiny{±0.8} &  \underline{93.7}\tiny{±0.3} & \textbf{74.3}\tiny{±0.8} & \textbf{84.3}\tiny{±0.7} & \textbf{79.0} (\textcolor{red}{{+3.7}}) \\
    \hline
    {full-FT} \cite{Zhang_2023_ICCV}  & DeiT \cite{deit}  &   90.0\tiny{±0.6}&92.5\tiny{±0.2}&65.3\tiny{±0.7}&89.8\tiny{±0.4}&73.9\tiny{±0.6}&83.3\tiny{±0.5}&70.3\tiny{±0.8}&92.2\tiny{±0.4}&83.0\tiny{±0.6}&85.0\tiny{±0.6}&82.5  \\ 
    \rowcolor{gray!12}   {{DETA}} \cite{Zhang_2023_ICCV} &   &   {\underline{90.8}\tiny{±0.6}} &  {\textbf{93.3\tiny{±0.2}}} &   {\textbf{71.6\tiny{±0.7}}} &   {\textbf{92.4\tiny{±0.4}}} &  {\underline{78.0}\tiny{±0.6}} &  {\underline{84.1}\tiny{±0.6}}  &  {\underline{75.2}\tiny{±0.8}} &   {\underline{84.4}\tiny{±0.4}}   &   {\textbf{95.5\tiny{±0.6}}} &  {\textbf{90.0\tiny{±0.6}}}  &  {\underline{85.2}}  (\textcolor{red}{{+2.7}})  \\
     \rowcolor{green!8} \textbf{{DETA}}++  &  &  \textcolor{black}{\textbf{91.4}\tiny{±0.6}}  & \underline{92.7}\tiny{±0.2}  &  \underline{70.9}\tiny{±0.7} & \textbf{92.4}\tiny{±0.4}  & \textbf{79.5}\tiny{±0.6}  & \textbf{85.0}\tiny{±0.6} & \textbf{75.5}\tiny{±0.8} & \textbf{88.6}\tiny{±0.4}  & \underline{94.1}\tiny{±0.6} & \underline{88.7}\tiny{±0.6} &  \textbf{85.9}  (\textcolor{red}{{+3.4}})\\
    \hline
    {full-FT}  \cite{Zhang_2023_ICCV}  &    SwinT \cite{liu2021swin}&90.8\tiny{±0.8}&91.2\tiny{±0.3}&57.6\tiny{±1.0}&88.3\tiny{±0.5}&76.4\tiny{±0.6}&81.9\tiny{±0.8}&67.8\tiny{±0.9}&92.3\tiny{±0.4}&82.5\tiny{±0.6}&83.9\tiny{±0.8}&81.3  \\ 
    \rowcolor{gray!12}  {{DETA}} \cite{Zhang_2023_ICCV} &  &      {\textbf{91.8\tiny{±0.9}}} &  {\textbf{92.5\tiny{±0.3}}} &   {\underline{68.9}\tiny{±0.9}} &   {\underline{92.7}\tiny{±0.5}} &  {\underline{79.5}\tiny{±0.7}} &  {\underline{82.8}\tiny{±0.6}}  &  {\underline{76.6}\tiny{±0.8}}  &   {\textbf{96.4\tiny{±0.4}}} &   {\underline{82.9}\tiny{±0.4}}   &  {\textbf{89.9\tiny{±0.7}}}  &  \underline{{85.4}} (\textcolor{red}{{+4.1}}) \\
    \rowcolor{green!8} \textbf{{DETA}}++   &  & \underline{91.6}\tiny{±1.0}   & \underline{92.3}\tiny{±0.3}  & \textbf{70.1}\tiny{±0.9}  & \textbf{93.6}\tiny{±0.5} & \textbf{80.9}\tiny{±0.6}  & \textbf{84.4}\tiny{±0.6} & \textbf{77.7}\tiny{±0.8}  & \underline{95.9}\tiny{±0.4}  & \textbf{88.4}\tiny{±0.4} & \underline{89.6}\tiny{±0.7} &  \textbf{86.5} (\textcolor{red}{{+5.2}})\\
    \hline
    \end{tabular}%
  \label{table1}%
\end{table*}%

\begin{table*}[htbp]
\setlength{\abovecaptionskip}{0.01cm}  
  \centering 
  \tabcolsep 0.1in
  \footnotesize
    \caption{Few-shot classification ACCs (\%) of six baseline methods w/ or w/o DETA \cite{Zhang_2023_ICCV} and DETA++ on the \textbf{OOD-polluted Meta-Dataset}. The results are the avg on the ten datasets of Meta-Dataset. The best and second best results on each baseline method are \textbf{bold}  and \underline{underlined}, respectively.}
    \begin{tabular}{l|l|ccccc}
    \hline
    \multicolumn{1}{l|}{\multirow{2}{*}{\makecell[c]{{Method}}}} & \multicolumn{1}{l|}{\multirow{2}{*}{\makecell[c]{Pret.Model}}} & \multicolumn{5}{c}{{Ratio of OOD noise} ($\alpha$)}\\
    \cline{3-7}  
    & & \multicolumn{1}{c}{0\%} & \multicolumn{1}{c}{10\%} & \multicolumn{1}{c}{30\%} & \multicolumn{1}{c}{50\%} & \multicolumn{1}{c}{70\%}  \\
    \hline
    {TSA} \cite{li2022cross}  & URL  \cite{liu2020universal} &75.5 & 71.4&64.4&56.0&36.7 \\
    \rowcolor{gray!12}    {{DETA}} \cite{Zhang_2023_ICCV} &  &  \underline{77.3} (\textcolor{red}{{+1.8}}) & \underline{73.7} (\textcolor{red}{{+2.3}})& \underline{66.5} (\textcolor{red}{{+2.1}})& \underline{57.9} (\textcolor{red}{{+1.9}}) & \underline{39.4} (\textcolor{red}{{+2.7}}) \\
     \rowcolor{green!8} \textbf{{DETA}}++ &  & \textbf{77.6} (\textcolor{red}{{+2.1}})  &  \textbf{74.1} (\textcolor{red}{{+2.7}})& \textbf{67.2} (\textcolor{red}{{+2.8}})& \textbf{58.7} (\textcolor{red}{{+2.7}}) & \textbf{40.6} (\textcolor{red}{{+3.9}}) \\
    \hline
    {eTT} \cite{xuexploring} & DINO \cite{caron2021emerging}& 80.9& 76.7&67.5&50.4&38.1\\
   \rowcolor{gray!12}    {{DETA}} \cite{Zhang_2023_ICCV}  &  &   {\underline{82.8}} (\textcolor{red}{{+1.9}}) & 
      \underline{79.4} (\textcolor{red}{{+2.7}})& \underline{69.9} (\textcolor{red}{{+2.4}})& \textbf{55.1} (\textcolor{red}{{+4.7}})& \underline{40.6} (\textcolor{red}{{+2.5}}) \\
      \rowcolor{green!8} \textbf{{DETA}}++ & & \textbf{83.6} (\textcolor{red}{{+2.7}})   &    \textbf{79.9} (\textcolor{red}{{+3.2}})& \textbf{71.2} (\textcolor{red}{{+3.7}})& \underline{54.3} (\textcolor{red}{{+3.9}}) & \textbf{42.8} (\textcolor{red}{{+4.7}}) \\
    \hline
      {full-FT} \cite{Zhang_2023_ICCV} & MoCo \cite{he2020momentum} &73.0 &66.9&62.8&54.1&34.8 \\
    \rowcolor{gray!12}  {{DETA}} \cite{Zhang_2023_ICCV} & &    {\underline{76.4}} (\textcolor{red}{{+3.4}}) & \underline{70.6} (\textcolor{red}{{+3.7}})& \underline{66.5} (\textcolor{red}{{+3.7}})& \underline{58.0} (\textcolor{red}{{+3.9}}) & \textbf{39.7} (\textcolor{red}{{+4.9}}) \\
     \rowcolor{green!8} \textbf{{DETA}}++ &  &  \textbf{77.8} (\textcolor{red}{{+4.8}}) &   \textbf{72.2} (\textcolor{red}{{+5.3}})& \textbf{68.0} (\textcolor{red}{{+5.2}})& \textbf{60.4} (\textcolor{red}{{+6.3}}) & \textbf{39.7} (\textcolor{red}{{+4.9}}) \\
    \hline 
    {full-FT} \cite{Zhang_2023_ICCV} & CLIP \cite{radford2021learning} &75.3 &70.1&64.9&54.0&35.7  \\
    \rowcolor{gray!12}  {{DETA}} \cite{Zhang_2023_ICCV} & &  {\underline{77.6}} (\textcolor{red}{{+2.3}}) &    \underline{73.0} (\textcolor{red}{{+2.9}})&  \underline{68.8} (\textcolor{red}{{+3.9}}) & \underline{56.6} (\textcolor{red}{{+2.6}}) & \underline{38.8} (\textcolor{red}{{+3.1}})\\
     \rowcolor{green!8} \textbf{{DETA}}++ & & \textbf{79.0} (\textcolor{red}{{+3.7}})  &    \textbf{74.2} (\textcolor{red}{{+4.1}})& \textbf{69.6} (\textcolor{red}{{+4.7}})& \textbf{58.4} (\textcolor{red}{{+4.4}}) & \textbf{41.0} (\textcolor{red}{{+5.3}}) \\
    \hline
    {full-FT} \cite{Zhang_2023_ICCV} & DeiT \cite{deit}& 82.5& 78.0 &75.4&63.8&43.2 \\
    \rowcolor{gray!12}  {{DETA}} \cite{Zhang_2023_ICCV} &  &  {\underline{85.2}}  (\textcolor{red}{{+2.7}})  &  \textbf{81.7} (\textcolor{red}{{+3.7}})& \underline{78.1} (\textcolor{red}{{+2.7}})&  \textbf{67.7} (\textcolor{red}{{+3.9}})& \underline{46.0} (\textcolor{red}{{+2.8}}) \\
     \rowcolor{green!8} \textbf{{DETA}}++ &  &  \textbf{85.9}  (\textcolor{red}{{+3.4}})  &   \underline{81.6} (\textcolor{red}{{+3.6}})& \textbf{79.4} (\textcolor{red}{{+4.0}})& \underline{67.3} (\textcolor{red}{{+3.5}}) & \textbf{47.7} (\textcolor{red}{{+3.5}}) \\
    \hline
    {full-FT} \cite{Zhang_2023_ICCV} & SwinT \cite{liu2021swin} &81.3 & 75.4&70.7&58.5&41.1  \\
  \rowcolor{gray!12}  {{DETA}} \cite{Zhang_2023_ICCV} &  &  \underline{{85.4}} (\textcolor{red}{{+4.1}}) &  \textcolor{black}{\underline{79.4} (\textcolor{black}{{+4.0}})} & \underline{76.0} (\textcolor{red}{{+5.3}})&  \textbf{{65.0}} (\textcolor{red}{{+6.5}})& \underline{46.7} (\textcolor{red}{{+5.6}})\\
    \rowcolor{green!8} \textbf{{DETA}}++ & &  \textbf{86.5} (\textcolor{red}{{+5.2}}) &    \textbf{80.9} (\textcolor{red}{{+5.5}})& \textbf{77.3} (\textcolor{red}{{+6.6}})& \underline{64.2} (\textcolor{red}{{+5.7}}) & \textbf{47.0} (\textcolor{red}{{+5.9}}) \\
    \hline
    \end{tabular} %
  \label{table2}%
\end{table*}%

\subsubsection{Experimental Setup}

\keypoint{Datasets}. 
We perform experiments on Meta-Dataset \cite{triantafillou2019meta}, a challenging FSL benchmark, which subsumes ten image datasets from various vision domains in one collection.
We consider two versions of Meta-Dataset in our experiments.
\begin{itemize}
\item \textit{Vanilla Meta-Dataset}: 
In the common setting of FSL, both support and query samples of few-shot tasks constructed from the vanilla Meta-Dataset are from ID classes (w/o OOD noise). Therefore, we can directly use the vanilla Meta-Dataset to verify the ID-denoising performance of DETA++.
\item \textit{OOD-polluted Meta-Dataset}:
We manually introduce different ratios (i.e., 10\%$\sim$70\%) of OOD noise (i.e., samples from unseen classes) to the ID classes of few-shot tasks constructed from the vanilla Meta-Dataset, based on which we can validate the OOD-denoising performance of DETA++.
\end{itemize} 

However, it is worth mentioning that in the standard task-sampling protocol for Meta-Dataset, the generated test tasks are way/shot imbalanced, {a.k.a.} \textit{varied-way varied-shot}. 
To avoid cases where the number of support samples in a class (i.e. shot) is less than 10, we adopt a unified task sampling protocol for the two Meta-Dataset versions by fixing the shot of every inner-task class to 10, {i.e.}, \textit{varied-way 10-shot} (in Table \ref{table1} and Table \ref{table2}). 
Yet, when conducting state-of-the-art comparison, we still employ the standard \textit{varied-way varied-shot} protocol for fair comparison (in Table \textcolor{magenta}{3}).

\begin{table*}[htbp]
\setlength{\abovecaptionskip}{0.01cm}  
  \centering
  \tabcolsep 0.022in
  \textcolor{black}{
   \caption{Comparison with state-of-the-art on Meta-Dataset. 
   $ ^{\clubsuit} $ the model is pre-trained on ImgNet; $ ^{\spadesuit} $ the model is jointly pre-trained on the eight datasets: ImgNet, Omglot, Acraft, CUB, DTD, QkDraw, Fungi and Flower.
   $\sharp$ the feature backbone is ViT-Tiny. The baseline of DETA++ is TSA \cite{li2022cross}. The best and second best results in each setting are \textbf{bold}  and \underline{underlined}, respectively. The ranks in () are used in the computation of the Friedman test \cite{friedman1937Friedman}.}}
    \begin{tabular}{l|llllllllll|l}
    \hline
    \multicolumn{1}{l|}{\multirow{2}{*}{\makecell[l]{Method$ ^{\clubsuit} $ (RN-18)}}} & \multicolumn{1}{c}{{In-Domain}}   & \multicolumn{9}{|c|}{{Out-of-Domain}} & \multirow{2}{*}{\textbf{Avg}} \\ 
    \cline{2-11} 
    \multicolumn{1}{l|}{} & \multicolumn{1}{l}{ImageNet} & \multicolumn{1}{|l}{Omglot} & \multicolumn{1}{l}{Acraft} & \multicolumn{1}{l}{CUB} & \multicolumn{1}{l}{DTD} & \multicolumn{1}{l}{QkDraw} & \multicolumn{1}{l}{Fungi} & \multicolumn{1}{l}{Flower} & \multicolumn{1}{l}{COCO} & \multicolumn{1}{l|}{Sign} & \\
    \hline
    Finetune  \cite{triantafillou2019meta}&45.8\tiny{±1.1} \scriptsize{(12)} &60.9\tiny{±1.6} \scriptsize{(11)} &68.7\tiny{±1.3} \scriptsize{(5)} &57.3\tiny{±1.3} \scriptsize{(11)} &69.0\tiny{±0.9} \scriptsize{(8)} &42.6\tiny{±1.2} \scriptsize{(12)} &38.2\tiny{±1.0} \scriptsize{(11)} &85.5\tiny{±0.7} \scriptsize{(9)} &34.9\tiny{±1.0} \scriptsize{(12)} &66.8\tiny{±1.3} \scriptsize{(5)} &57.0 \scriptsize{(9.6)} \\
    ProtoNet  \cite{triantafillou2019meta} &50.5\tiny{±1.1} \scriptsize{(9)} &60.0\tiny{±1.4} \scriptsize{(12)} &53.1\tiny{±1.0} \scriptsize{(9)} &68.8\tiny{±1.0} \scriptsize{(9)} &66.6\tiny{±0.8} \scriptsize{(10)} &49.0\tiny{±1.1} \scriptsize{(11)} &39.7\tiny{±1.1} \scriptsize{(10)} &85.3\tiny{±0.8} \scriptsize{(10.5)} &41.0\tiny{±1.1}\scriptsize{(11)} &47.1\tiny{±1.1} \scriptsize{(11)} &56.1 \scriptsize{(10.25)} \\
    FoProMA \cite{triantafillou2019meta}&49.5\tiny{±1.1} \scriptsize{(10)} &63.4\tiny{±1.3} \scriptsize{(8)} &56.0\tiny{±1.0} \scriptsize{(7)} &68.7\tiny{±1.0} \scriptsize{(10)} &66.5\tiny{±0.8} \scriptsize{(11)} &51.5\tiny{±1.0} \scriptsize{(9)} &40.0\tiny{±1.1} \scriptsize{(9)} &87.2\tiny{±0.7} \scriptsize{(6)} &43.7\tiny{±1.1} \scriptsize{(9)} &48.8\tiny{±1.1} \scriptsize{(9)} &57.5 \scriptsize{(8.8)} \\
    Alfa-FoPro.  \cite{triantafillou2019meta}&52.8\tiny{±1.1} \scriptsize{(7)} &61.9\tiny{±1.5} \scriptsize{(9)} &63.4\tiny{±1.1} \scriptsize{(6)} &69.8\tiny{±1.1} \scriptsize{(8)} &70.8\tiny{±0.9} \scriptsize{(7)} &59.2\tiny{±1.2} \scriptsize{(5)} &41.5\tiny{±1.2} \scriptsize{(6)} &86.0\tiny{±0.8} \scriptsize{(8)} &48.1\tiny{±1.1} \scriptsize{(7)} &60.8\tiny{±1.3} \scriptsize{(6)} &61.4 \scriptsize{(6.9)} \\
    BOHB  \cite{saikia2020optimized}&51.9\tiny{±1.1} \scriptsize{(8)} &67.6\tiny{±1.2} \scriptsize{(7)} &54.1\tiny{±0.9} \scriptsize{(8)} &70.7\tiny{±0.9} \scriptsize{(7)} &68.3\tiny{±0.8} \scriptsize{(9)} &50.3\tiny{±1.0} \scriptsize{(10)} &41.4\tiny{±1.1} \scriptsize{(7)} &87.3\tiny{±0.6} \scriptsize{(5)} &48.0\tiny{±1.0} \scriptsize{(8)} &51.8\tiny{±1.0} \scriptsize{(8)} &59.1 \scriptsize{(7.7)} \\
    FLUTE \cite{triantafillou2021learning}  &46.9\tiny{±1.1} \scriptsize{(11)} &61.6\tiny{±1.4} \scriptsize{(10)} &48.5\tiny{±1.0} \scriptsize{(12)} &47.9\tiny{±1.0} \scriptsize{(12)} &63.8\tiny{±0.8} \scriptsize{(12)} &57.5\tiny{±1.0} \scriptsize{(6)} &31.8\tiny{±1.0} \scriptsize{(12)} &80.1\tiny{±0.9} \scriptsize{(12)} &41.4\tiny{±1.0} \scriptsize{(10)} &46.5\tiny{±1.1} \scriptsize{(12)} &52.6 \scriptsize{(10.9)} \\
    eTT$^{\sharp}$ \cite{xuexploring} &56.4\tiny{±1.1} \scriptsize{(6)} &72.5\tiny{±1.4} \scriptsize{(6)} &72.8\tiny{±1.0} \scriptsize{(3)} &73.8\tiny{±1.1} \scriptsize{(4)} &77.6\tiny{±0.8} \scriptsize{(3)} &68.0\tiny{±0.9} \scriptsize{(3)} &\textbf{51.2}\tiny{±1.1} \scriptsize{(1)} &\textbf{93.3}\tiny{±0.6} \scriptsize{(1)} &55.7\tiny{±1.0} \scriptsize{(4)} &84.1\tiny{±1.0} \scriptsize{(3)} &70.5 \scriptsize{(3.4)} \\
    {URL} \cite{li2021universal}  &56.8\tiny{±1.0} \scriptsize{(5)} &79.5\tiny{±0.8} \scriptsize{(4)} &49.4\tiny{±0.8} \scriptsize{(11)} &71.8\tiny{±0.9} \scriptsize{(6)} &72.7\tiny{±0.7} \scriptsize{(6)} &53.4\tiny{±1.0} \scriptsize{(8)} &40.9\tiny{±0.9} \scriptsize{(8)} &85.3\tiny{±0.7} \scriptsize{(10.5)} &52.6\tiny{±0.9} \scriptsize{(6)} &47.3\tiny{±1.0} \scriptsize{(10)} &61.1 \scriptsize{(7.45)} \\
    {Beta} \cite{li2021universal} &58.4\tiny{±1.1} \scriptsize{(4)} &{\underline{81.1}\tiny{±0.8}} \scriptsize{(2)} &51.9\tiny{±0.9} \scriptsize{(10)} &73.6\tiny{±1.0} \scriptsize{(5)} &74.0\tiny{±0.7} \scriptsize{(5)} &55.6\tiny{±1.0} \scriptsize{(7)} &42.2\tiny{±0.9} \scriptsize{(5)} &86.2\tiny{±0.8} \scriptsize{(7)} &55.1\tiny{±1.0} \scriptsize{(5)} &59.0\tiny{±1.1} \scriptsize{(7)} &63.7 \scriptsize{(5.7)} \\
    {TSA} \cite{li2022cross} &59.5\tiny{±1.1} \scriptsize{(3)} &78.2\tiny{±1.2} \scriptsize{(5)} &72.2\tiny{±1.0} \scriptsize{(4)} &74.9\tiny{±0.9} \scriptsize{(3)} &77.3\tiny{±0.7} \scriptsize{(4)} &67.6\tiny{±0.9} \scriptsize{(4)} &44.7\tiny{±1.0} \scriptsize{(4)} &90.9\tiny{±0.6} \scriptsize{(4)} &59.0\tiny{±1.0} \scriptsize{(3)} &82.5\tiny{±0.8} \scriptsize{(4)} &70.7 \scriptsize{(3.8)} \\  
    \rowcolor{gray!12}  {{DETA}} \cite{Zhang_2023_ICCV} (Ours) & {\underline{60.7}\tiny{±1.0}} \scriptsize{(2)} &\textbf{{81.6}}\tiny{±1.1} \scriptsize{(1)} &{\underline{73.0}\tiny{±1.0}} \scriptsize{(2)} &{\underline{77.0}\tiny{±0.9}} \scriptsize{(2)} &{\underline{78.3}\tiny{±0.7}} \scriptsize{(2)} &{\underline{69.5}\tiny{±0.9}} \scriptsize{(2)} &{\underline{47.6}\tiny{±1.0}} \scriptsize{(2)} &{{92.6\tiny{±0.6}}} \scriptsize{(3)} &{\underline{60.3}\tiny{±1.0}} \scriptsize{(2)} & {\underline{86.8}\tiny{±0.8}} \scriptsize{(2)} &\underline{{72.8}} \scriptsize{(2.0)} \\
    \rowcolor{green!8} \textbf{{DETA}}++ (Ours)  & \textbf{{61.8}}\tiny{±1.0} \scriptsize{(1)} &{81.0\tiny{±1.2}} \scriptsize{(3)} &\textbf{{74.7}}\tiny{±1.0} \scriptsize{(1)} &\textbf{{77.9}}\tiny{±0.9} \scriptsize{(1)} &\textbf{{79.6}}\tiny{±0.7} \scriptsize{(1)} &\textbf{{71.0}}\tiny{±0.9} \scriptsize{(1)} &{{47.4\tiny{±1.0}}} \scriptsize{(3)} &{\underline{92.9}\tiny{±0.6}} \scriptsize{(2)} &\textbf{{61.7}}\tiny{±1.0} \scriptsize{(1)} &\textbf{{87.4}}\tiny{±0.9} \scriptsize{(1)} &\textbf{{73.6}} \scriptsize{(1.5)} \\
    \hline \hline
    \multicolumn{1}{l|}{\multirow{2}{*}{\makecell[c]{Method$ ^{\spadesuit} $ (RN-18)}}} & \multicolumn{8}{c}{{In-Domain}}   & \multicolumn{2}{|c|}{{Out-of-Domain}} & \multirow{2}{*}{\textbf{Avg}} \\ 
    \cline{2-11}
    \multicolumn{1}{c|}{} & \multicolumn{1}{l}{ImageNet} & \multicolumn{1}{l}{Omglot} & \multicolumn{1}{l}{Acraft} & \multicolumn{1}{l}{CUB} & \multicolumn{1}{l}{DTD} & \multicolumn{1}{l}{QkDraw} & \multicolumn{1}{l}{Fungi} & \multicolumn{1}{l}{Flower} & \multicolumn{1}{|l}{COCO} & \multicolumn{1}{l|}{Sign} &  \\
    \hline
    CNAPS  \cite{requeima2019fast} &50.8\tiny{±1.1} \scriptsize{(12)} &91.7\tiny{±0.5} \scriptsize{(11)} &83.7\tiny{±0.6} \scriptsize{(10)} &73.6\tiny{±0.9} \scriptsize{(11)} &59.5\tiny{±0.7} \scriptsize{(12)} &74.7\tiny{±0.8} \scriptsize{(12)} &50.2\tiny{±1.1} \scriptsize{(9)} &88.9\tiny{±0.5} \scriptsize{(10)} &39.4\tiny{±1.0} \scriptsize{(12)} &56.5\tiny{±1.1} \scriptsize{(9)} &66.9 \scriptsize{(10.8)} \\
    SimpCNAPS  \cite{bateni2020improved} &58.4\tiny{±1.1} \scriptsize{(6)} &91.6\tiny{±0.6} \scriptsize{(12)} &82.0\tiny{±0.7} \scriptsize{(12)} &74.8\tiny{±0.9} \scriptsize{(10)} &68.8\tiny{±0.9} \scriptsize{(9.5)} &76.5\tiny{±0.8} \scriptsize{(11)} &46.6\tiny{±1.0} \scriptsize{(12)} &90.5\tiny{±0.5} \scriptsize{(8.5)} &48.9\tiny{±1.1} \scriptsize{(9)} &57.2\tiny{±1.0} \scriptsize{(8)} &69.5 \scriptsize{(9.8)} \\
    TransCNAPS \cite{bateni2022enhancing} &57.9\tiny{±1.1} \scriptsize{(7)} &94.3\tiny{±0.4} \scriptsize{(6)} &84.7\tiny{±0.5} \scriptsize{(9)} &78.8\tiny{±0.7} \scriptsize{(7)} &66.2\tiny{±0.8} \scriptsize{(11)} &77.9\tiny{±0.6} \scriptsize{(9)} &48.9\tiny{±1.2} \scriptsize{(10)} &92.3\tiny{±0.4} \scriptsize{(3)} &42.5\tiny{±1.1} \scriptsize{(11)} &59.7\tiny{±1.1} \scriptsize{(6)} &70.3 \scriptsize{(7.9)} \\
    SUR \cite{dvornik2020selecting} &56.2\tiny{±1.0} \scriptsize{(10)} &94.1\tiny{±0.4} \scriptsize{(8)} &85.5\tiny{±0.5} \scriptsize{(8)} &71.0\tiny{±1.0} \scriptsize{(12)} &71.0\tiny{±0.8} \scriptsize{(8)} &81.8\tiny{±0.6} \scriptsize{(5.5)} &64.3\tiny{±0.9} \scriptsize{(6)} &82.9\tiny{±0.8} \scriptsize{(12)} &52.0\tiny{±1.1} \scriptsize{(7)} &51.0\tiny{±1.1} \scriptsize{(11)} &71.0 \scriptsize{(8.75)} \\
    URT  \cite{liu2020universal} &56.8\tiny{±1.1} \scriptsize{(9)} &94.2\tiny{±0.4} \scriptsize{(7)} &85.8\tiny{±0.5} \scriptsize{(7)} &76.2\tiny{±0.8} \scriptsize{(8)} &71.6\tiny{±0.7} \scriptsize{(6)} &82.4\tiny{±0.6} \scriptsize{(4)} &64.0\tiny{±1.0} \scriptsize{(7)} &87.9\tiny{±0.6} \scriptsize{(11)} &48.2\tiny{±1.1} \scriptsize{(10)} &51.5\tiny{±1.1} \scriptsize{(10)} &71.9 \scriptsize{(7.9)} \\
    FLUTE \cite{triantafillou2021learning} &58.6\tiny{±1.0} \scriptsize{(5)} &92.0\tiny{±0.6} \scriptsize{(10)} &82.8\tiny{±0.7} \scriptsize{(11)} &75.3\tiny{±0.8} \scriptsize{(9)} &71.2\tiny{±0.8} \scriptsize{(7)} &77.3\tiny{±0.7} \scriptsize{(10)} &48.5\tiny{±1.0} \scriptsize{(11)} &90.5\tiny{±0.5} \scriptsize{(8.5)} &52.8\tiny{±1.1} \scriptsize{(6)} &63.0\tiny{±1.0} \scriptsize{(5)} &71.2 \scriptsize{(8.25)} \\
    Tri-M  \cite{liu2021multi} &51.8\tiny{±1.1} \scriptsize{(11)} &93.2\tiny{±0.5} \scriptsize{(9)} &87.2\tiny{±0.5} \scriptsize{(6)} &79.2\tiny{±0.8} \scriptsize{(6)} &68.8\tiny{±0.8} \scriptsize{(9.5)} &79.5\tiny{±0.7} \scriptsize{(8)} &58.1\tiny{±1.1} \scriptsize{(8)} &91.6\tiny{±0.6} \scriptsize{(6)} &50.0\tiny{±1.0} \scriptsize{(8)} &58.4\tiny{±1.1} \scriptsize{(7)} &71.8 \scriptsize{(7.85)} \\
    {URL}  \cite{li2021universal}  &57.0\tiny{±1.0} \scriptsize{(8)} &94.4\tiny{±0.4} \scriptsize{(5)} &88.0\tiny{±0.5} \scriptsize{(5)} &80.3\tiny{±0.7} \scriptsize{(5)} &74.6\tiny{±0.7} \scriptsize{(5)} &81.8\tiny{±0.6} \scriptsize{(5.5)} &66.2\tiny{±0.9} \scriptsize{(5)} &91.5\tiny{±0.5} \scriptsize{(7)} &54.1\tiny{±1.0} \scriptsize{(5)} &49.8\tiny{±1.0} \scriptsize{(12)} &73.8 \scriptsize{(6.25)} \\
    {Beta}  \cite{li2021universal} &58.8\tiny{±1.1} \scriptsize{(4)} &94.5\tiny{±0.4} \scriptsize{(4)} &89.4\tiny{±0.4} \scriptsize{(4)} &80.7\tiny{±0.8} \scriptsize{(4)} &77.2\tiny{±0.7} \scriptsize{(4)} &82.5\tiny{±0.6} \scriptsize{(3)} &68.1\tiny{±0.9} \scriptsize{(3)} &92.0\tiny{±0.5} \scriptsize{(5)} &57.3\tiny{±1.0} \scriptsize{(4)} &63.3\tiny{±1.1} \scriptsize{(4)} &76.4 \scriptsize{(3.9)} \\
    {TSA}   \cite{li2022cross} &59.5\tiny{±1.0} \scriptsize{(3)} &94.9\tiny{±0.4} \scriptsize{(3)} &89.9\tiny{±0.4} \scriptsize{(3)} &81.1\tiny{±0.8} \scriptsize{(3)} &77.5\tiny{±0.7} \scriptsize{(3)} &81.7\tiny{±0.6} \scriptsize{(7)} &66.3\tiny{±0.8} \scriptsize{(4)} &92.2\tiny{±0.5} \scriptsize{(4)} &57.6\tiny{±1.0} \scriptsize{(3)} &82.8\tiny{±1.0} \scriptsize{(3)} &78.3 \scriptsize{(3.6)} \\
    \rowcolor{gray!12}  {{DETA}} \cite{Zhang_2023_ICCV} (Ours) &{\underline{61.0}\tiny{±1.0}} \scriptsize{(2)} &{\underline{95.6}\tiny{±0.4}} \scriptsize{(2)} &\textbf{{91.4}}\tiny{±0.4} \scriptsize{(1)} &{\underline{82.7}\tiny{±0.7}} \scriptsize{(2)} &{\underline{78.9}}\tiny{±0.7} \scriptsize{(2)} &{\underline{83.4}\tiny{±0.6}} \scriptsize{(2)} &{\underline{68.2}\tiny{±0.8}} \scriptsize{(2)} &{\textbf{93.4}\tiny{±0.5}} \scriptsize{(1)} &{\underline{58.5}\tiny{±1.0}} \scriptsize{(2)} &{\underline{86.9}\tiny{±1.0}} \scriptsize{(2)} &\underline{{80.1}} \scriptsize{(1.8)} \\
    \rowcolor{green!8} \textbf{{DETA}}++ (Ours) &\textbf{{63.1}}\tiny{±1.0} \scriptsize{(1)} &\textbf{{96.4}}\tiny{±0.5} \scriptsize{(1)} &{\underline{91.3}\tiny{±0.4}} \scriptsize{(2)} &\textbf{{84.8}\tiny{±0.7}} \scriptsize{(1)} &{\textbf{79.8}\tiny{±0.7}} \scriptsize{(1)} &{\textbf{84.3}\tiny{±0.6}} \scriptsize{(1)} &{\textbf{70.2}\tiny{±0.8}} \scriptsize{(1)} &{\underline{92.9}\tiny{±0.5}} \scriptsize{(2)} &{\textbf{59.7}\tiny{±1.0}} \scriptsize{(1)} &{\textbf{88.0}\tiny{±0.9}} \scriptsize{(1)} &\textbf{{81.2}} \scriptsize{(1.2)} \\
    \hline
    \end{tabular}
    \label{table:sota2}
\end{table*}%

\keypoint{Baselines.}
We demonstrate the effectiveness and flexibility of DETA++ by applying it to a broad spectrum of baseline methods applied on various  diverse pre-trained models.  
For adapter-FT methods, we consider the two strong baselines {TSA} \cite{li2022cross} and {eTT} \cite{xuexploring}.
TSA integrates a residual adapter to the single-domain URL (w/ ResNet-18 pre-trained on $84\times84$ ImageNet) \cite{li2021universal} and eTT attaches a self-attention adapter to DINO (w/ ViT-Small) \cite{caron2021emerging}\footnote{\url{https://github.com/facebookresearch/dino}}.
For full-finetuning (full-FT) methods, motivated by \cite{li2022cross}, 
we compute the training loss and perform inference with a nonparametric nearest centroid classifier (NCC) instead of a linear classifier that is commonly-used in transfer learning literature. 
We perform full-FT to adapt various pre-trained models, including MoCo (w/ ResNet-50) \cite{he2020momentum}\footnote{\url{https://github.com/facebookresearch/moco}}, CLIP (w/ ResNet-50) \cite{radford2021learning}\footnote{\url{https://github.com/OpenAI/CLIP}}, DeiT (w/ ViT-Small) \cite{deit}\footnote{\url{https://github.com/facebookresearch/deit}} and the vanilla Swin Transformer (SwinT) \cite{liu2021swin}\footnote{\url{https://github.com/microsoft/Swin-Transformer}}. 
All models are trained on Imagenet-1k, except for CLIP, which is trained on large-scale image-text pairs.
For each baseline, we match the image size in model pre-training and task adaptation, {i.e.}, the image size is set to $84\times84$ for {TSA} \cite{li2022cross}, and $224\times224$ for others.

\keypoint{Evaluation Metric.}
We evaluate the few-shot classification performance of different methods on 600 randomly sampled test tasks for each dataset from the Meta-Dataset, and report average accuracy (ACC, in \%) and 95\% confidence intervals.

\keypoint{Implementation Details.}
We perform our experiments on the Meta-Dataset benchmark \cite{triantafillou2019meta} based on the open-source  repository of TSA \cite{li2022cross}\footnote{\url{https://github.com/VICO-UoE/URL}}. 
Specifically, we perform task adaptation by updating the parameters of the pre-trained model or the appended task adapter for 40 iterations on the target task.
\textcolor{black}{All images in each dataset are first re-scaled to $\Psi \times \Psi$, based on the model architectures of baselines, i.e., $\Psi = 84$ for TSA \cite{li2022cross} and $\Psi = 224$ for other methods.}
During each iteration of DETA++, 4 and 2 local regions are cropped from every image for TSA and other methods, respectively. 
The projection head is a two-layer MLP, and we follow \cite{khosla2020supervised} to set the embedding dimension to 128. 
For fair comparisons, DETA++ uses the same learning rate $\gamma$ for each baseline method during task adaptation.
For the adapter-FT methods {TSA} and {eTT} \cite{xuexploring}, we set $\gamma$ to 0.05 and 0.001, respectively; for the full-FT methods, denoted as full-FT\&MoCo, full-FT\&CLIP, full-FT\&DeiT, and full-FT\&SwinT, we set $\gamma$ to 0.001, 0.01, 0.1, 0.05, respectively.
We adjust the coefficient $\beta$, the region size $\Omega$, and the threshold $\varrho$ in ablation studies. The implementation code is based on PyTorch. The experiments are performed on a V100 GPU.

\begin{table*}
\setlength{\abovecaptionskip}{0.01cm}  
		\centering
		 \tabcolsep 0.029in
		\footnotesize
       \textcolor{black}{\caption{Few-shot classification performance of DETA++ with different designs. The ratio of OOD noise in support samples is 30\%.  The results of each baseline method are averaged over the ten datasets of Meta-Dataset. Idx-A = Baseline, Idx-E = DETA++.}}
        \begin{tabular}{c|ccccc|cc|cc|cc|cc|cc|cc|cc}
			\hline
			\multicolumn{1}{l|}{\multirow{2}{*}{\makecell[l]{Idx}}} &\multicolumn{5}{c|}{\textbf{Setting of DETA++} (Ours)} &  \multicolumn{2}{c|}{\textbf{Avg}} &  \multicolumn{2}{c|}{URL \cite{li2021universal} } &  \multicolumn{2}{c|}{DINO \cite{caron2021emerging}}  &  \multicolumn{2}{c|}{MoCo \cite{he2020momentum} } &  \multicolumn{2}{c|}{CLIP \cite{radford2021learning}} &  \multicolumn{2}{c|}{DeiT \cite{deit}}  &  \multicolumn{2}{c}{SwinT \cite{liu2021swin}} \\
			\cline{2-20}
			&CoRA & $\mathcal{L}_{{clean}}$  & $\mathcal{L}_{{noise}}$ & IntraSwap  & LocalNCC &  ID-de. &  OOD-de. &  ID- &  OOD- & ID- &  OOD- & ID- &  OOD-&  ID- &  OOD-&  ID- &  OOD-&  ID- &  OOD- \\
			\hline
			A &\textcolor{gray!50}{\XSolidBrush} & \textcolor{gray!50}{\XSolidBrush} & \textcolor{gray!50}{\XSolidBrush} & \textcolor{gray!50}{\XSolidBrush} & \textcolor{gray!50}{\XSolidBrush} &78.0&67.6&75.5 &64.4&80.9&67.5& 73.0 & 62.8 &75.3&64.9&82.5&75.4&81.3&70.7\\
        \hline
    		B &	\Checkmark &  \Checkmark & \textcolor{gray!50}{\XSolidBrush} & \textcolor{gray!50}{\XSolidBrush} & \textcolor{gray!50}{\XSolidBrush}   
            &79.8&69.2&76.3&64.8&82.3&68.8&74.5   &  64.0 & 77.1   & 66.5 &84.1&77.6&84.2&73.4 \\
		C&	\Checkmark &  \Checkmark & \textcolor{gray!50}{\XSolidBrush} & \textcolor{gray!50}{\XSolidBrush} & \Checkmark 
            &80.4&70.2&77.0&66.1&82.6&69.4&75.4 & 64.5 & 77.2 & 67.9 &84.7&77.5&85.7&76.0 \\
		D&	\Checkmark &  \Checkmark&\Checkmark  & \textcolor{gray!50}{\XSolidBrush} & \Checkmark  &81.2&71.6&77.4 &66.8 &83.3&70.6 &76.8 &66.7 &78.6 &69.5  &85.5&79.0&86.1&76.7 \\
        \hline
		\rowcolor{green!8} E	&\Checkmark &  \Checkmark&\Checkmark  &\Checkmark  &\Checkmark  &\textbf{81.7} &\textbf{72.1}  &\textbf{77.6}&\textbf{67.2}&\textbf{83.6}&\textbf{71.2}&\textbf{77.8}  &\textbf{68.0} & \textbf{79.0}& \textbf{69.6} &\textbf{85.9}&\textbf{79.4}&\textbf{86.5}&\textbf{77.3}\\
        \hline  \hline
        \multicolumn{6}{c|}{\textbf{DETA}\cite{Zhang_2023_ICCV}}   &80.7&70.9&77.3 &66.5&82.8&69.9 &76.4 &66.5 &77.6 &68.8&85.2&78.1&85.4&76.0 \\
            \hline
		\end{tabular}
		\label{ab}
\end{table*}

\subsubsection{ID-denoising Performance}
The few-shot classification results of the six baseline approaches w/ or w/o DETA \cite{Zhang_2023_ICCV} and DETA++ on the vanilla Meta-Dataset are reported in Table \ref{table1}. 
We have the following observations from the obtained results.
\textbf{\underline{Firstly}}, DETA++ consistently improves adapter-FT and full-FT based task adaptation methods, which confirms that DETA++ is orthogonal to those methods and able to improve model robustness to ID noise.
\textcolor{black}{In particular, we observe that DeiT and SwinT. networks pre-trained with ImageNet class labels have a specific advantage, this is probably due the class overlap between ImageNet and COCO class labels.}
\textbf{\underline{Secondly}}, DETA++ achieves significant performance gains on both TSA (for $84\times84$-size input images) and other methods (for $224\times224$-size images), suggesting DETA++ is model-agnostic.
\textbf{\underline{Thirdly}}, DETA++ successfully overcomes two types of ID noise: {background clutter} (on ImageNet, CUB, and etc.) and {image corruption} (on Omglot and QkDraw), qualitative results are shown in Fig. \ref{vis1} and discussed in Section \ref{sectionvis}.
\textbf{\underline{Finally}}, on the six baselines, DETA++ yields superior performance than DETA, further demonstrating the effectiveness and flexibility of the proposed DETA++ framework on ID-denoising.
Importantly, the achieved results uncover the ever-overlooked ID noise in the  Meta-Dataset benchmark. More qualitative evidence for this problem are discussed in Section \ref{sectionvis} and demonstrated in Fig. \ref{vis1}.

\subsubsection{OOD-denoising Performance}
We further demonstrate the effectiveness of DETA++ by performing OOD-denoising on the OOD-polluted Meta-Dataset.
Concretely, we manually introduce different ratios (10\%$\sim$70\%) of OOD samples in support samples for each task.
Table \ref{table2} reports the average ACC of different baselines methods w/ or w/o DETA \cite{Zhang_2023_ICCV} and DETA++ on the ten Meta-Dataset datasets, under different ratios of OOD noise.
We have the following observations.
\textbf{\underline{Firstly}}, the few-shot classification performance gradually decreases as the ratio of OOD noise in support samples increases.
\textbf{\underline{Secondly}}, both DETA and DETA++ consistently improves the baseline methods by a large margin across all settings, demonstrating their effectiveness to improve model robustness to OOD noise.
\textbf{\underline{Thirdly}}, compared with the obtained ID-denoising results in Table \ref{table1}, the OOD-denoising performance of DETA and DETA++ are more significant.
Possible reasons are twofold. 
\textbf{i)} The negative impact of OOD noise on performance is more significant than that of ID noise, as the OOD noise/samples contain almost no valuable object features associated with the ID classes.
\textbf{ii)} When one class contains samples from other classes, our designed CoRA can identify the harmful regions more precisely by taking advantage of out-of-class relevance information.
\textbf{\underline{Finally}}, DETA++ consistently outperforms the strong competitor DETA on the six baselines, demonstrating the effectiveness and flexibility of the proposed DETA++ on OOD-denoising.

\subsubsection{Comparison with State-of-the-art}
\label{friedman}
So far, we can see that our DETA++ can be flexibly plugged into both adapter-finetuning and full-finetuning based task adaptation methods to improve model robustness to ID  and OOD noises.
It is interesting to investigate whether DETA++ can further boost the current state-of-the-art after tackling the ID noise in the vanilla  Meta-Dataset.
Hence, we apply our DETA++ to the strong baseline scheme TSA\cite{li2022cross} and conduct experiments on  Meta-Dataset with a group of competitors, e.g., FLUTE\cite{triantafillou2021learning}, URL\cite{li2021universal}, eTT\cite{xuexploring} and DETA \cite{Zhang_2023_ICCV}. 
In Table \textcolor{magenta}{3}, we can observe that DETA++ considerably improves the strong baseline TSA and establishes new state-of-the-art results on nearly all ten Meta-Dataset datasets. 
Besides, DETA++ outperforms the strong competitor DETA in both in-domain and out-of-domain generalization settings, which further confirms the effectiveness and flexibility of our proposed DETA++ framework. 

\textcolor{black}{Additionally, we conduct the Friedman test \cite{friedman1937use,garcia2010advanced} on the results presented in Table \textcolor{magenta}{3} to facilitate more rigorous comparisons with those state-of-the-art methods. The Friedman test is a non-parametric statistical method used to determine whether there are significant differences in the performance of multiple methods across multiple datasets (please refer to \cite{friedman1937use,garcia2010advanced} for more details).
Table \textcolor{magenta}{3} (Top) reports the accuracy ranks of $k=12$ different methods on $n=10$ datasets. Therefore, 
$$\chi_{F}^{2}=\frac{12 n}{k(k+1)}\left[\sum_{j=1}^k (\frac{1}{n}\sum_{i=1}^nr_i^j)^{2}-\frac{k(k+1)^{2}}{4}\right]=87.396,$$ $F_{F}=\frac{(n-1) \chi_{F}^{2}}{n(k-1)-\chi_{F}^{2}}=34.798$. Similarly, for Table \textcolor{magenta}{3} (Down), $\chi_{F}^{2}=83.585$, $F_{F}=28.478$. For Table \textcolor{magenta}{3} (Top)/(Down), $F_{F}$ is distributed according to the $F$ distribution with $k-1=11$ and $(k-1)(n-1)=99$. The $p$ value computed by using the $F(11,99)$ distribution is  $<10^{-5}$, so the null hypothesis is rejected at a high level of significance. 
This further demonstrates the effectiveness of our proposed DETA++.}

\subsubsection{Ablation Studies}
\label{ablation}
In this section, we conduct ablative studies to scrutinize our DETA++ approach. 
\textcolor{black}{
Unless stated otherwise, the ratio of OOD noise in support samples is 30\%, and the average ACC on the ten datasets of Meta-Dataset are reported.}

\keypoint{Effectiveness of the Designed Components.}
\textcolor{black}{The proposed DETA++ framework consists of the following components:
\textbf{i}) a CoRA module;
\textbf{ii}) a clean prototype loss $\mathcal{L}_{clean}$;
\textbf{iii}) a noise entropy maximization loss $\mathcal{L}_{noise}$;
\textbf{iv}) an intra-class region swapping (IntraSwap) scheme;
and \textbf{v}) a local nearest centroid classifier (LocalNCC).
We perform a component-wise analysis to examine the impact of each component by sequentially adding them to the six baseline methods. 
The obtained results are reported in Table \textcolor{magenta}{4}, where each baseline method is equipped with a nearest centroid classifier (NCC).
We have the following observations from the table.
\textcolor{black}{\textbf{\underline{Firstly}}, each component of DETA++ contributes to performance enhancement across the six baseline methods. }
\textbf{\underline{Secondly}}, guided by the CoRA module, the two losses $\mathcal{L}_{clean}$ and $\mathcal{L}_{noise}$ complement each other to improve the ID- and OOD-denoising performance, simultaneously. In particular, DETA++ outperforms DETA \cite{Zhang_2023_ICCV} by 0.5\% in ID-denoising and 0.7\% in OOD-denoising, even without integrating IntraSwap. This underscores the superiority of DETA++’s losses $\mathcal{L}_{clean}$ and $\mathcal{L}_{noise}$ over DETA’s losses $\mathcal{L}_{l}$ and $\mathcal{L}_{g}$ in tackling the dual noises in support samples. 
\textbf{\underline{Thirdly}}, by further exploring the “free-lunch” in the CoRA module (i.e., the computed weights of images and image regions of support samples), the developed training-time IntraSwap and test-time LocalNCC boost the performance of DETA++.
\textbf{\underline{Finally}}, by combining all those components, DETA++ outperforms the baseline methods by large margins. On average, the combination of all components of DETA++ achieves significant improvements of 3.7\% in ID-denoising and 4.5\% in OOD-denoising, surpassing DETA’s improvements of 2.7\% and 3.3\%, respectively.
In a nutshell, the achieved results in the table demonstrates the effectiveness of DETA++ in overcoming the adverse effects of ID and OOD noise in support samples of few-shot tasks.}

\keypoint{Impact of the Coefficient $\beta$.}
In the learning objective of our DETA++ (Eq. \ref{eq12}), we adopt the coefficient $\beta$ to control the contribution of $\mathcal{L}_{noise}$.
In this experiment, we carefully tune $\beta$ by setting it to the values of \{0.01, 0.05, 0.1, 0.3, 0.7, 1.0\}, and report the average ACC on the ten datasets of Meta-Dataset in Table \ref{hypers}.
As can be observed, our DETA++ is not sensitive to the change of $\beta$ within a certain range (from 0.05 to 0.7).
Besides, DETA++ outperforms the baseline significantly, particularly when the value of $\beta$ falls within the range of 0.1 to 0.7.
DETA++ achieves the best ID- and OOD-denoising results when $\beta=0.3$, simultaneously.

\keypoint{Impact of the Threshold $\varrho$.}
The proposed DETA++ leverages a threshold $\varrho$ to filter out noisy regions and images according to their weights computed by the CoRA module.
In this experiment, we adjust $\varrho$ by setting it to \{0.05, 0.1, 0.3, 0.5, 0.7, 1.0\}, and report the average ACC on the ten datasets of Meta-Dataset in Table \ref{hypers}.
As can be observed, our DETA++ achieves relatively stable results and outperforms the baseline significantly when $\varrho$ takes the values of 0.3 and 0.5.
Particularly, our DETA++ yields the best ID- and OOD-denoising results when $\varrho$ is set to 0.3.

\begin{table}
\setlength{\abovecaptionskip}{0.01cm}  
		\centering
		 \tabcolsep 0.10in
		\footnotesize
        \caption{Impacts of the coefficient $\beta$ and the threshold $\varrho$ on performance. The baseline is full-FT\&MoCo  \cite{he2020momentum}, the ratio of OOD noise in support samples is 30\%, and the average ACC on the ten datasets of Meta-Dataset are reported.}
        \begin{tabular}{c|llllll}
			\hline
		     $\beta$ & 0.01  & 0.05 & 0.1  & 0.3  & 0.7& 1.0 \\
			\hline
			ID-denoising & 75.5  &  76.2 & 77.4 &  \textbf{77.8} & 76.6 &  75.8 \\
			OOD-denoising & 63.5  &  66.8 & 67.5 & \textbf{68.0}  & 67.7 &  67.2 \\
			\hline \hline
		    $\varrho$ & 0.05  & 0.1 & 0.3  & 0.5  & 0.7& 1.0 \\
			\hline
			ID-denoising & 74.5  &  76.4 &  \textbf{77.8}  & 77.1 & 76.8 &  75.4 \\
			OOD-denoising & 64.1  &  67.3 & \textbf{68.0} & 67.2  & 67.5 &  65.5 \\
            \hline
		\end{tabular}
		\label{hypers}
\end{table}

\begin{table}
\setlength{\abovecaptionskip}{0.01cm}  
		\centering
		 \tabcolsep 0.10in
		\footnotesize
      \textcolor{black}{\caption{Impacts of the region size $\Omega$ on performance. The baseline for $84\times 84$ images is TSA \cite{li2022cross}, and for $224\times 224$ images is full-FT\&MoCo \cite{he2020momentum}. The ratio of OOD noise in support samples is 30\%, and the average ACC on the ten datasets of Meta-Dataset are reported.}}  
        \begin{tabular}{l|llllll}
			\hline
		     $\Omega$ ($84\times 84$) & 32  & 40 & 48  &  56  & 64 & 72 \\
			\hline
			ID-denoising & 75.6  &  76.2 & 77.1 & \textbf{77.6} & {77.0} &  76.3 \\
			OOD-denoising & 64.8  &  65.7 & 66.3 & \textbf{67.2}  & 66.5 &  66.0 \\
			\hline \hline
		     $\Omega$ ($224\times 224$)& 48  & 64 & 84  &  112  & 128 & 168 \\
			\hline
			ID-denoising & 74.0  &  75.3 & 76.9 & {77.8} & \textbf{78.0} &  76.7 \\
			OOD-denoising & 63.2  &  64.6 & 67.1 & \textbf{68.0}  & 67.5 &  66.9 \\
            \hline
		\end{tabular}
		\label{hypers2}
\end{table}

\keypoint{\textcolor{black}{Impact of the {{Region Size}} $\Omega$.}} \textcolor{black}{
In our experiments, all images from each dataset are first re-scaled to $\Psi \times \Psi$, based on the model architectures of baseline methods, e.g., $\Psi = 84$ for TSA \cite{li2022cross} and $\Psi = 224$ for full-FT\&MoCo \cite{he2020momentum}. During each training iteration of DETA++, the devised CoRA module crops a set of $\Omega\times \Omega$ image regions from the input $\Psi \times \Psi$ images and determine the weights of these image regions. 
In this part, we scrutinize the impact of $\Omega$ on the performance of DETA++.
Specifically, we compare the results of DETA++ when $\Omega$ takes the values of \{32, 40, 48, 56, 64, 72\} and \{48, 64, 84, 112, 128, 168\} for $84\times 84$ and $224\times 224$ images, respectively. 
The results are shown in Table \textcolor{magenta}{6}, where the baseline for $84\times 84$ images is TSA, and for $224\times 224$ images is full-FT\&MoCo.
From the results, we have the following observations.
\textbf{\underline{Firstly}}, when the size of the cropped image region is too small, the performance improvement of DETA++ over the baseline methods becomes less significant. 
One possible reason is that cropped image regions with excessively small sizes struggle to capture object relevant features.
\textbf{\underline{Secondly}}, DETA++ is generally insensitive to changes in $\Omega$ when $\Omega$ takes the values of \{48, 56, 64\} for $84\times 84$ images and \{84, 112, 128\} for $224\times 224$ images. 
\textbf{\underline{Finally}},  DETA++ establishes the optimal results when $\Omega=56$ and $112$ for $84\times 84$ and $224\times 224$ images, respectively.}

\begin{figure} 
\setlength{\abovecaptionskip}{-0.01cm}  
	\centering
	\includegraphics[width=1\linewidth]{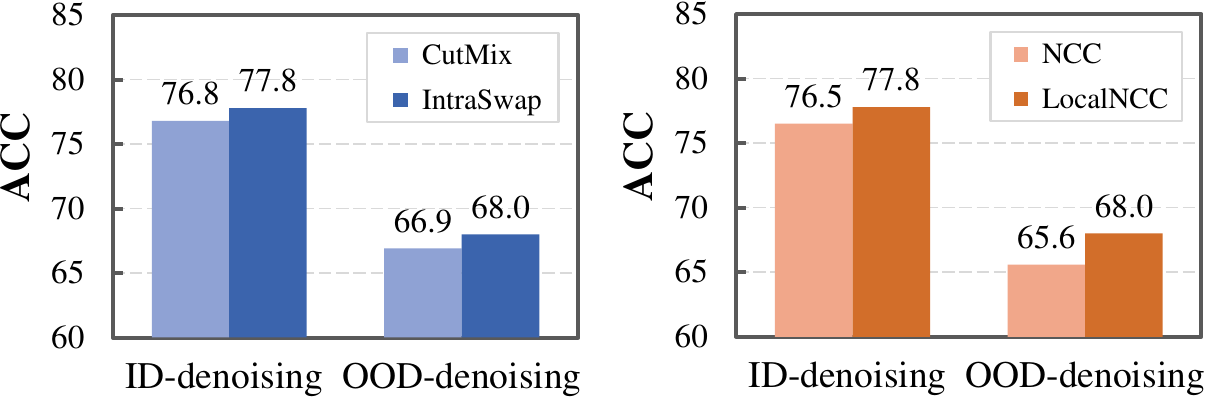}
	\caption{Effectiveness of our designed IntraSwap and LocalNCC in our DETA++. The baseline is full-FT\&MoCo, the ratio of OOD noise in support samples is 30\%, and the average ACC on the ten datasets of Meta-Dataset are reported.}
		\label{figcub}  
\end{figure}

\begin{table}
\setlength{\abovecaptionskip}{-0.01cm}  
		\centering
		 \tabcolsep 0.03in
		\footnotesize
      \textcolor{black}{\caption{Computational cost of DETA \cite{Zhang_2023_ICCV} and DETA++, w.r.t. training and inference time. The ratio of OOD noise in support samples is 30\%, and the average ACC on the ten datasets of Meta-Dataset are reported. Experiments are performed on a V100 GPU.}}
        \begin{tabular}{l|c|cc|cc}
			\hline
			\multicolumn{1}{l|}{\multirow{2}{*}{\makecell[l]{Method}}} & \multicolumn{1}{l|}{\multirow{2}{*}{\makecell[l]{Pret.Model}}} & Train. & Infer. & \multicolumn{2}{c}{ACC} \\
			\cline{5-6}
			&   & (s/iter.) & (ms/img) & ID-denois.& OOD-denois. \\
			\hline
            TSA\cite{li2022cross} &URL\cite{liu2020universal} & 0.11  & 8  & 75.5 & 64.4 \\
		\rowcolor{gray!12} 	DETA \cite{Zhang_2023_ICCV}& & 0.15 &  8  & 77.3(\textcolor{red}{+1.8})  &66.5 (\textcolor{red}{+2.1})  \\
		\rowcolor{green!8} 	\textbf{DETA}++& & 0.18 & 9  &\textbf{77.6} (\textcolor{red}{+2.1})  &\textbf{67.2} (\textcolor{red}{+2.8}) \\
            \hline
        {eTT} \cite{xuexploring} & DINO \cite{caron2021emerging} & 0.14  & 11  & 80.9 & 67.5 \\
		\rowcolor{gray!12} 	DETA \cite{Zhang_2023_ICCV} && 0.16 &  13  & 82.8 (\textcolor{red}{+1.9})  &69.9 (\textcolor{red}{+2.4})  \\
		\rowcolor{green!8} 	\textbf{DETA}++ & &0.18 & 13  &\textbf{83.6} (\textcolor{red}{+2.7})  &\textbf{71.2} (\textcolor{red}{+3.7}) \\
            \hline
            full-FT \cite{Zhang_2023_ICCV} &MoCo \cite{he2020momentum} & 0.17  & 15  & 73.0 & 62.8 \\
		\rowcolor{gray!12} 	DETA \cite{Zhang_2023_ICCV}& & 0.25 &  15  & 76.4 (\textcolor{red}{+3.4})  &66.5 (\textcolor{red}{+3.7})  \\
		\rowcolor{green!8} 	\textbf{DETA}++ & &0.28 & 16  &\textbf{77.8} (\textcolor{red}{+4.8})  &\textbf{68.0} (\textcolor{red}{+5.2}) \\
            \hline
        full-FT \cite{Zhang_2023_ICCV} &CLIP \cite{radford2021learning}& 0.22  & 21  & 75.3 & 64.9 \\
		\rowcolor{gray!12} 	DETA \cite{Zhang_2023_ICCV}& & 0.27 &  22  & 77.6 (\textcolor{red}{+2.3})  &68.8 (\textcolor{red}{+3.9})  \\
		\rowcolor{green!8} 	\textbf{DETA}++ && 0.29 & 22  &\textbf{79.0} (\textcolor{red}{+3.7})  &\textbf{69.6} (\textcolor{red}{+4.7}) \\
            \hline
        full-FT \cite{Zhang_2023_ICCV} &DeiT \cite{deit} & 0.19  & 13  & 82.5 & 75.4 \\
		\rowcolor{gray!12} 	DETA \cite{Zhang_2023_ICCV} && 0.24 &  15 & 85.2 (\textcolor{red}{+2.7})  &78.1 (\textcolor{red}{+2.7})  \\
		\rowcolor{green!8} 	\textbf{DETA}++ & &0.28 & 15  &\textbf{85.9} (\textcolor{red}{+3.4})  &\textbf{79.4} (\textcolor{red}{+4.0}) \\
            \hline
        full-FT \cite{Zhang_2023_ICCV}&SwinT \cite{liu2021swin} & 0.25  & 22  & 81.3 & 70.7 \\
		\rowcolor{gray!12} 	DETA \cite{Zhang_2023_ICCV}& & 0.33 &  23  & 85.4 (\textcolor{red}{+4.1})  &76.0 (\textcolor{red}{+5.3})  \\
		\rowcolor{green!8} 	\textbf{DETA}++ && 0.35 & 24  &\textbf{86.5} (\textcolor{red}{+5.2})  &\textbf{77.3} (\textcolor{red}{+6.6}) \\
            \hline
		\end{tabular}
  \label{table.time}
\end{table}

\keypoint{IntraSwap vs. CutMix \cite{yun2019cutmix}.}
To rectify the prototypes of ID classes in Eq. (\ref{eq.proto}), DETA++ further explores the “free-lunch” in the CoRA module to establish a parameter-free distribution augmentation scheme, named IntraSwap.
Unlike CutMix \cite{yun2019cutmix} that generates a new training image by cutting and pasting a random image region from one image onto another, IntraSwap mixes the clean regions stored in the memory bank with one image from the same class to prevent the loss of valuable information in the image.
We report the ID- and OOD-denoising results of our DETA++ framework w/ IntraSwap or CutMix in Fig. \ref{figcub} (\textit{Left}), where IntraSwap shows superior performance than CutMix.

\keypoint{LocalNCC vs. NCC \cite{li2021universal}.}
Unlike DETA \cite{Zhang_2023_ICCV} that uses global images to build a nearest centroid classifier (NCC) in a high-dimensional representation space, DETA++ leverages clean regions of each class stored in the memory bank to construct a LocalNCC in a learned low-dimensional embedding space for noise-robust inference.
To demonstrate the superiority of our LocalNCC over NCC, we present the ID- and OOD-denoising results of DETA++ using either LocalNCC or NCC for inference, as shown in Fig. \ref{figcub} (\textit{Right}).
As can be observed, LocalNCC outperforms NCC by a large margin, demonstrating LocalNCC's ability of eliminating the adverse effect of ID and OOD noise in few-shot tasks.

\subsubsection{Computational Cost}
\textcolor{black}{So far, we have demonstrated the effectiveness and flexibility of DETA++ in improving the few-shot classification performance under the dual noises.
This raises a key question: how much additional computational cost does DETA++ incur while achieving such performance improvements on those methods?
To answer this question, we report the training time (denoted as: seconds per iteration) and inference time (denoted as: millisecond per image) of the six baseline methods integrated with or without DETA \cite{Zhang_2023_ICCV}  and DETA++ in Table \textcolor{magenta}{7}. 
As we can see, compared to the six baseline methods, both DETA and DETA++ incur higher training time cost while maintaining comparable inference time cost.
DETA and DETA++ consistently outperform the six baselines, with DETA++ achieving particularly remarkable results.}
In a nutshell, due to the much smaller number of samples in few-shot tasks, the computational cost of DETA and DETA++ is not significant and may be acceptable in the vast majority of realistic scenarios. 
This also suggests that further improving the computational efficiency of DETA and DETA++ on more generic scenarios with a huge number of training data may be the key to boosting their practicability in future work.

\subsection{Evaluation on Few-shot OOD Detection}
\label{fsood}
\textcolor{black}{
Few-shot OOD detection aims to detect out-of-distribution (OOD) samples that the model has not been exposed to during training using only a few labeled support samples of the target task. The (few-shot) OOD detection performance of a learned model serves as a direct indicator of its ability to distinguish ID images from OOD noise in query samples at inference. Importantly, enhanced OOD detection results not only reflect improved separability between ID and OOD distributions but also suggest that the model has learned discriminative characteristics for ID classes, reducing the adverse impact of ID noise in query samples (such as cluttered backgrounds from ID images) at inference. 
In this part, we compare the OOD detection performance of models learned by DETA++ against those trained with other approaches to validate the effectiveness of DETA++ in addressing the dual noises in query samples.}


Recent prompt tuning based task adaptation methods have demonstrated significant improvements in few-shot OOD detection. Hence, we apply DETA++ to the prompt tuning baseline method CoOp \cite{zhou2022coop}, which performs task adaptation on the pre-trained CLIP \cite{radford2021learning} model using support samples of the target task\footnote{Please refer to CoOp \cite{zhou2022coop} or our prior works \cite{zhangji2023dept,wu2024skip} for more details about CLIP based prompt tuning.}. 
Note that during test-time OOD detection (i.e., inference), DETA++ uses a MCM \cite{ming2022delving} head rather the LocalNCC (devised for few-shot classification) to classify testing images to ID and OOD samples. 
In other words, DETA++ only retains the four components CoRA, $\mathcal{L}_{clean}$, $\mathcal{L}_{noise}$ and IntraSwap for prompt tuning based few-shot OOD detection. 
The ablation studies on the four components are presented in Section \ref{oodablation}.

\begin{table*}[]
\centering
\setlength{\abovecaptionskip}{0.01cm}  
\tabcolsep 0.021in
\textcolor{black}{
\caption{Comparison with state-of-the-art on the ImageNet-1K OOD detection benchmark. The baseline of DETA \cite{Zhang_2023_ICCV} and DETA++ is {CoOp$_{\mathrm{MCM}}$ \cite{miyai2024locoop}}. \\
$ ^{\downarrow} $ smaller is better; $^{\uparrow} $ larger is better. The best and second best results on  are \textbf{bold} and \underline{underlined}. 
All methods are based on ViT or CLIP models, with the exception of SCALE \cite{xu2023scaling}, which uses the ResNet-50 architecture.
The ranks in () are used in the computation of the Friedman test \cite{friedman1937Friedman}.}}
\begin{tabular}{l|llllllllll|ll}
\hline
{\multirow{2}{*}{\textbf{Method}}} & \multicolumn{2}{c}{iNaturalist}      & \multicolumn{2}{c}{SUN} & \multicolumn{2}{c}{Places365} & \multicolumn{2}{c}{Texture} &
\multicolumn{2}{c|}{\textcolor{black}{NINCO}} & \multicolumn{2}{c}{\textbf{Avg}}\\
       & \textbf{FPR$\downarrow$} & \textbf{AUROC$\uparrow$} & \textbf{FPR$\downarrow$} & \textbf{AUROC$\uparrow$} & \textbf{FPR$\downarrow$} & \textbf{AUROC$\uparrow$} & \textbf{FPR$\downarrow$} & \textbf{AUROC$\uparrow$} & \textcolor{black}{\textbf{FPR$\downarrow$}} & \textcolor{black}{\textbf{AUROC$\uparrow$}}& \textbf{FPR$\downarrow$} & \textbf{AUROC$\uparrow$} \\
\hline
{MCM \cite{ming2022delving}}  &30.94 \scriptsize{(13)} &94.61 \scriptsize{(10)} &37.67 \scriptsize{(9)} &92.56 \scriptsize{(8)} &44.76 \scriptsize{(11)} &89.76 \scriptsize{(8)} &57.91 \scriptsize{(13)} &86.10 \scriptsize{(13)} &\textcolor{black}{66.01 \scriptsize{(14)}} &\textcolor{black}{76.95 \scriptsize{(13)}} &47.46 \scriptsize{(12)} &88.00 \scriptsize{(10.4)} \\
{GL-MCM \cite{miyai2023glmcm}}  &15.18 \scriptsize{(3)} &96.71 \scriptsize{(3)} &30.42 \scriptsize{(4)} &93.09 \scriptsize{(4)} &38.85 \scriptsize{(5)} &89.90 \scriptsize{(7)} &57.93 \scriptsize{(14)} &83.63 \scriptsize{(15)} &\textcolor{black}{57.33 \scriptsize{(9)}} &\textcolor{black}{81.18 \scriptsize{(10)} }&39.94 \scriptsize{(7)} &88.90 \scriptsize{(7.8)} \\
\textcolor{black}{SCALE \cite{xu2023scaling}}  &\textcolor{black}{31.56 \scriptsize{(14)}} &\textcolor{black}{91.85 \scriptsize{(15)}} &\textcolor{black}{41.51 \scriptsize{(11)}} &\textcolor{black}{90.73 \scriptsize{(10)}} &\textcolor{black}{53.37 \scriptsize{(13)}} &\textcolor{black}{84.90 \scriptsize{(14)}} &\textcolor{black}{53.41 \scriptsize{(11)}} &\textcolor{black}{87.29 \scriptsize{(9)}} &{\textcolor{black}{61.22 \scriptsize{(11)}}} &{\textcolor{black}{78.14 \scriptsize{(12)}}} &\textcolor{black}{48.21 \scriptsize{(12)}} &\textcolor{black}{86.58 \scriptsize{(12)}} \\
\textcolor{black}{NNGuide \cite{park2023nearest}}  &\textcolor{black}{27.61 \scriptsize{(9)}} &\textcolor{black}{93.22 \scriptsize{(13)}} &\textcolor{black}{45.29 \scriptsize{(13)}} &\textcolor{black}{89.01 \scriptsize{(13)}} &\textcolor{black}{40.13 \scriptsize{(8)}} &\textcolor{black}{89.49 \scriptsize{(10)}} &\textcolor{black}{51.07 \scriptsize{(9)}} &\textcolor{black}{87.13 \scriptsize{(11)}} &\textcolor{black}{54.78 \scriptsize{(7)}} &\textcolor{black}{81.83 \scriptsize{(8)}} &\textcolor{black}{43.78 \scriptsize{(9.2)}} &\textcolor{black}{88.14 \scriptsize{(11)}} \\
{ODIN \cite{liang2018odin}}  &30.22 \scriptsize{(12)} &94.65 \scriptsize{(9)} &54.04 \scriptsize{(15)} &87.17 \scriptsize{(15)} &55.06 \scriptsize{(14)} &85.54 \scriptsize{(13)} &51.67 \scriptsize{(10)} &87.85 \scriptsize{(8)} &\textcolor{black}{62.82 \scriptsize{(12)}} &\textcolor{black}{78.43 \scriptsize{(11)}}&50.76 \scriptsize{(12.6)} &86.73 \scriptsize{(11.2)} \\
{ViM \cite{wang2022vim}}   &32.19 \scriptsize{(15)} &93.16 \scriptsize{(14)} &54.01 \scriptsize{(14)} &87.19 \scriptsize{(14)} &60.67 \scriptsize{(15)} &83.75 \scriptsize{(15)} &53.94 \scriptsize{(12)} &87.18 \scriptsize{(10)} &\textcolor{black}{65.48 \scriptsize{(13)}} &\textcolor{black}{71.29 \scriptsize{(15)}} &53.26 \scriptsize{(13.8)} &84.51 \scriptsize{(13.6)} \\
{KNN  \cite{sun2022knn}}   &29.17 \scriptsize{(11)} &94.52 \scriptsize{(11)} &35.62 \scriptsize{(6)} &92.67 \scriptsize{(6)} &39.61 \scriptsize{(6)} &91.02 \scriptsize{(4)} &64.35 \scriptsize{(15)} &85.67 \scriptsize{(14)} &\textcolor{black}{71.83 \scriptsize{(15)}} &\textcolor{black}{74.60 \scriptsize{(14)}} &48.12 \scriptsize{(10.6)} &87.70 \scriptsize{(9.8)} \\
{NPOS$_{\mathrm{MCM}}$ \cite{tao2022npos}}  &16.58 \scriptsize{(5)} &96.19 \scriptsize{(5)} &43.77 \scriptsize{(12)} &90.44 \scriptsize{(11)} &45.27 \scriptsize{(12)} &89.44 \scriptsize{(11)} &46.12 \scriptsize{(7)} &88.80 \scriptsize{(6)} &\textcolor{black}{55.72 \scriptsize{(8)}} &\textcolor{black}{84.54 \scriptsize{(6)}} &41.49 \scriptsize{(8.8)} &89.88 \scriptsize{(7.8)} \\
{NPOS$_{\mathrm{GL}}$ \cite{tao2022npos}} &18.70 \scriptsize{(6)} &95.36 \scriptsize{(7)} &38.99 \scriptsize{(10)} &90.33 \scriptsize{(12)} &41.86 \scriptsize{(9)} &89.36 \scriptsize{(12)} &47.89 \scriptsize{(8)} &86.44 \scriptsize{(12)} &\textcolor{black}{58.19 \scriptsize{(10)}} &\textcolor{black}{83.46 \scriptsize{(7)}} &41.13 \scriptsize{(8.6)} &88.99 \scriptsize{(10)} \\
\hline
{CoOp$_{\mathrm{MCM}}$ \cite{miyai2024locoop}} &28.00 \scriptsize{(10)} &94.43 \scriptsize{(12)} &36.95 \scriptsize{(8)} &92.29 \scriptsize{(9)} &43.03 \scriptsize{(10)} &89.74 \scriptsize{(9)} &\textbf{39.33} \scriptsize{(1)} &\underline{91.24} \scriptsize{(2)} &\textcolor{black}{50.11 \scriptsize{(3)}} &\textcolor{black}{\underline{86.92} \scriptsize{(2)}} &39.48 \scriptsize{(6.4)} &90.92 \scriptsize{(6.8)} \\
{CoOp$_{\mathrm{GL}}$ \cite{miyai2024locoop}} &\underline{14.60} \scriptsize{(2)} &96.62 \scriptsize{(4)} &28.48 \scriptsize{(3)} &92.65 \scriptsize{(7)} &36.49 \scriptsize{(4)} &89.98 \scriptsize{(6)} &43.13 \scriptsize{(5)} &88.03 \scriptsize{(7)} &\textcolor{black}{52.34 \scriptsize{(4)}} &\textcolor{black}{81.51 \scriptsize{(9)}} &35.01 \scriptsize{(3.6)} &89.76 \scriptsize{(6.6)} \\
{LoCoOp$_{\mathrm{MCM}}$ \cite{miyai2024locoop}} &23.06 \scriptsize{(7)} &95.45 \scriptsize{(6)} &32.70 \scriptsize{(5)} &93.35 \scriptsize{(3)} &39.92 \scriptsize{(7)} &90.64 \scriptsize{(5)} &\underline{40.23} \scriptsize{(2)} &\textbf{91.32} \scriptsize{(1)} &\textcolor{black}{\underline{49.92} \scriptsize{(2)}} &\textcolor{black}{86.89 \scriptsize{(3)}} &37.17 \scriptsize{(4.6)} &91.53 \scriptsize{(3.6)} \\
{LoCoOp$_{\mathrm{GL}}$ \cite{miyai2024locoop}} &16.05 \scriptsize{(4)} &\underline{96.86} \scriptsize{(2)} &\underline{23.44} \scriptsize{(2)} &\textbf{95.07} \scriptsize{(1)} &{32.87} \scriptsize{(3)} &{91.98} \scriptsize{(3)} &42.28 \scriptsize{(4)} &90.19 \scriptsize{(5)} &\textcolor{black}{52.80 \scriptsize{(6)} }&\textcolor{black}{85.57 \scriptsize{(4)}} &\underline{33.49} \scriptsize{(3.8)} &\underline{91.93} \scriptsize{(3)} \\
\rowcolor{gray!12} {\textbf{DETA}} \cite{Zhang_2023_ICCV} (Ours)  &{25.58} \scriptsize{(8)} &{94.72} \scriptsize{(8)} &{36.64} \scriptsize{(7)} &{92.81} \scriptsize{(5)} &\underline{31.47} \scriptsize{(2)} &\underline{92.13} \scriptsize{(2)} &43.41 \scriptsize{(6)} &{90.28} \scriptsize{(4)} &\textcolor{black}{52.61 \scriptsize{(5)} }&\textcolor{black}{85.25 \scriptsize{(5)}} &37.94 \scriptsize{(5.6)} &91.04 \scriptsize{(4.8)} \\
\rowcolor{green!8} {\textbf{DETA++}}  (Ours)  &\textbf{14.04} \scriptsize{(1)} &\textbf{97.71} \scriptsize{(1)} &\textbf{21.52} \scriptsize{(1)} &\underline{94.73} \scriptsize{(2)} &\textbf{28.95} \scriptsize{(1)} &\textbf{92.60} \scriptsize{(1)} &41.61 \scriptsize{(3)} &{91.17} \scriptsize{(3)} &\textcolor{black}{\textbf{48.27} \scriptsize{(1)}} &\textcolor{black}{\textbf{88.15} \scriptsize{(1)} }&\textbf{30.88} \scriptsize{(1.4)} &\textbf{92.87} \scriptsize{(1.6)} \\
\hline
\end{tabular}
\label{tableood}
\end{table*}

\subsubsection{Experimental Setup}
\keypoint{Datasets}.
\textcolor{black}{
Following the common splits of the ImageNet-1K OOD detection benchmark \cite{huang2021mos},  we use ImageNet-1K \cite{deng2009imagenet} as the ID dataset (with 16 samples from each class serving as support samples) and use iNaturalist \cite{van2018inaturalist}, SUN \cite{xiao2010sun}, Places365 \cite{zhou2017places}, and Texture \cite{cimpoi2014textures} as OOD datasets. 
Despite the widespread use of this benchmark, previous work \cite{bitterwolf2023or} has revealed class overlap between the ID dataset and the four OOD datasets. Thus, we also include NINCO \cite{bitterwolf2023or}, an improved OOD dataset without overlapping classes with ImageNet-1k, for comprehensive evaluation.}

\keypoint{Evaluation Metric}.
We follow the widely employed setup in OOD detection to use the following evaluation metrics:
{FPR} (a.k.a. FPR95) \cite{macedo2021entropic}: The false positive rate of OOD samples when the true positive rate of ID examples reaches 95\%; \textit{smaller is better}.
{AUROC} \cite{hendrycks2016baseline}: The area under the curve of the receiver operating characteristic; \textit{larger is better}.
Note that we don't need to manually tune the threshold $\varepsilon$ for FPR and AUROC at inference, as the two metrics can determine $\varepsilon$ according to the classification results of testing ID samples.

\keypoint{Implementation Details.}
We adopt CoOp$_{\mathrm{MCM}}$ \cite{miyai2024locoop} as the baseline, which uses the publicly available CLIP-ViT-B/16 model as vision/text encoder. 
It is worth noting that CoRA directly uses the $14\times 14$ local features extracted by the vision encoder of CLIP to compute the weight of each region, i.e., there is no region cropping operation on the input images. 
The class prototypes in Eq. (\ref{eq.proto}) are the text embeddings of class names. 
The generated images by IntraSwap are used to augment the training data, rather than refining the class prototypes as in few-shot classification.
Besides, we use the loss function of DETA++ in Eq. (\ref{eq12}) to perform CLIP-based prompt tuning on the support samples of the ID dataset ImageNet-1K. The hyper-parameter $\beta$ is set to 1.0. We follow the few-shot evaluation protocol in CoOp \cite{zhou2022coop} and LoCoOp \cite{miyai2024locoop} to set the number of support samples per class to 16 for training.
The training epoch, learning rate and batch size take the values of 50, 0.01, and 16, respectively. 
Our implementation is based on the code of LoCoOp \cite{miyai2024locoop}\footnote{\url{https://github.com/AtsuMiyai/LoCoOp}.}.

\subsubsection{Comparison with State-of-the-art}
{To assess DETA++'s effectiveness, we compare its few-shot OOD detection performance with all comparison methods from \cite{miyai2024locoop}, i.e., zero-shot OOD detection methods MCM \cite{ming2022delving} and GL-MCM \cite{miyai2023glmcm}; full-finetuning based  methods ODIN \cite{liang2018odin}, ViM \cite{wang2022vim}, KNN \cite{sun2022knn} and NPOS$_{\mathrm{MCM}}$ \cite{tao2022npos}; and prompt learning based methods CoOp$_{\mathrm{MCM}}$ \cite{miyai2024locoop} and LoCoOp$_{\mathrm{MCM}}$ \cite{miyai2024locoop}. 
\textcolor{black}{
Beyond these methods from \cite{miyai2024locoop}, our comparisons also incorporate two recently proposed zero-shot OOD detection methods, SCALE \cite{xu2023scaling} and NNGuide \cite{park2023nearest}.
The results for the two methods are obtained using their default experimental settings.}}
To validate DETA++'s improvements over its predecessor DETA, we compare DETA++ with DETA \cite{Zhang_2023_ICCV} by applying DETA to the same baseline CoOp$_{\mathrm{MCM}}$ \cite{miyai2024locoop} and using a MCM \cite{ming2022delving} head to produce the OOD detection results of the adapted model. 
The obtained results on the OOD detection benchmark ImageNet-1K are reported in Table \textcolor{magenta}{8}, where we have the following observations.
\textbf{\underline{Firstly}}, prompt tuning based methods in general achieve superior performance than zero-shot and  full-finetuning based methods.
\textcolor{black}{
\textbf{\underline{Secondly}}, DETA++ shows superior performance than those strong competitors across the five OOD datasets, and establishes new state-of-the-art results. Notably, DETA++ reduces the average
FPR95 of the baseline method CoOp$_{\mathrm{MCM}}$ \cite{miyai2024locoop} by {{8.6}}\%.
\textbf{\underline{Thirdly}}, DETA++ outperforms DETA by a significant margin, proving the effectiveness of $\mathcal{L}_{noise}$ that leverages the detected noisy regions of the CoRA module to enhance the decision boundary between class features and ID/OOD noise features.
\textbf{\underline{Fourthly}}, the vast majority of methods perform less effectively on the NINCO dataset compared to the other four datasets, indicating that the NINCO dataset is more challenging.}
\textbf{\underline{Finally}}, DETA++ and DETA obtain inferior performance to both CoOp$_{\mathrm{MCM}}$ and LoCoOp$_{\mathrm{MCM}}$ on the Texture dataset. 
This is probably due to that the global and local features of Texture images are consistent and clean enough—there is almost no noisy background regions; as a result, DETA++ and DETA cannot fully show its advantages in this scenario. In a nutshell, the proposed DETA++ framework can function as a plugin to enhance the robustness of prompt tuning or task adaptation-based few-shot OOD detection methods against dual noises.


\textcolor{black}{Furthermore, we perform the Friedman test \cite{friedman1937use,garcia2010advanced} on the results to enable more rigorous comparisons with state-of-the-art methods. 
As shown, Table \textcolor{magenta}{8} reports the FPR (AUROC) ranks of $k=13$ different methods on $n=5$ datasets. 
According to Section \ref{friedman}, for the FPR (AUROC) ranks: [$\chi_{F}^{2}=49.360$, $F_{F}=9.565$] ([$\chi_{F}^{2}=44.660$, $F_{F}=7.049$]) is distributed according to the $F$ distribution with $k-1=12$ and $(k-1)(n-1)=48$. 
The $p$ value for both the FPR and AUROC ranks computed by using the $F(12,48)$ distribution is $<10^{-5}$, so the null hypothesis is rejected at a high level of significance. 
This further validates the effectiveness of our proposed DETA++.
}

\begin{table}[tp]
\setlength{\abovecaptionskip}{0.01cm}  
		\centering
		 \tabcolsep 0.042in
		\footnotesize
        \caption{\textcolor{black}{OOD detection performance of DETA++ with different designs. During inference, the devised LocalNCC for few-shot classification is replaced by a widely-used MCM \cite{ming2022delving} head for few-shot OOD detection. The baseline is {CoOp$_{\mathrm{MCM}}$ \cite{miyai2024locoop}}, the average results on the five OOD datasets are reported.  $ ^{\downarrow} $ smaller is better; $^{\uparrow} $ larger is better.}}
        \begin{tabular}{ccccc|cc}
			\hline
			\multicolumn{5}{c|}{Setting}  & \multicolumn{2}{c}{Avg} \\
			\cline{1-7}
			CoRA & $\mathcal{L}_{clean}$  & $\mathcal{L}_{noise}$ & IntraSwap &MCM \cite{ming2022delving} &\textbf{FPR$\downarrow$} & \textbf{AUROC$\uparrow$} \\
			\hline
			 \textcolor{gray!50}{\XSolidBrush} & \textcolor{gray!50}{\XSolidBrush} &  \textcolor{gray!50}{\XSolidBrush} & \textcolor{gray!50}{\XSolidBrush}  & \Checkmark & 39.48 & 90.92 \\
			\Checkmark &  \Checkmark & \textcolor{gray!50}{\XSolidBrush} & \textcolor{gray!50}{\XSolidBrush}  & \Checkmark & 36.34 &91.47 \\
   			\Checkmark &  \Checkmark&\Checkmark  &  \textcolor{gray!50}{\XSolidBrush} & \Checkmark  & {32.16}& {92.41}\\
		\rowcolor{green!8} 	\Checkmark &  \Checkmark&\Checkmark  &  \Checkmark & \Checkmark & \textbf{30.88}  & \textbf{92.87}   \\
            \hline
		\end{tabular}
		\label{abood}
\end{table}

\begin{table*}[htbp]
\setlength{\abovecaptionskip}{0.01cm}  
  \centering 
  \tabcolsep 0.13in
  \footnotesize
    \textcolor{black}{\caption{Quantitative comparison of the devised losses in DETA \cite{Zhang_2023_ICCV} and DETA++, in terms of absolute performance gains (\%) over baselines. The results are averaged over 11 datasets and 6 baselines (i.e., URL, DINO, MoCO, CLIP, DeiT and SwinT) for few-shot classification, and over 5 datasets and 1 baseline (i.e., CoOp$_{\mathrm{MCM}}$) for few-shot OOD detection. The ratio of OOD noise in support samples is 30\% for OOD-denoising. }}
    \begin{tabular}{l|cc|cc|cc|cc}
    \hline
    \multicolumn{1}{l|}{\multirow{2}{*}{\makecell[c]{{Setting}}}} & \multicolumn{2}{c|}{DETA losses} & \multicolumn{2}{c|}{DETA++ losses} & \multicolumn{2}{c|}{Few-shot Classification} & \multicolumn{2}{c}{Few-shot OOD Detection}\\
    \cline{2-9}  
    & $\mathcal{L}_l$ & $\mathcal{L}_g$  & $\mathcal{L}_{clean}$  & $\mathcal{L}_{noise}$  & ID-deno.{{$\uparrow$}}   &  OOD-deno.{{$\uparrow$}} & \quad  {\textbf{FPR$\downarrow$}} & {\textbf{AUROC$\uparrow$}}  \\
    \hline
    DETA++ w/ $\mathcal{L}_l$ \& $\mathcal{L}_g$ &\Checkmark&\Checkmark&\textcolor{gray!50}{\XSolidBrush}&\textcolor{gray!50}{\XSolidBrush}&+2.9&+3.5&\quad \textcolor{black}{-3.72}&+0.47\\
    DETA++ w/ $\mathcal{L}_g$ \& $\mathcal{L}_{noise}$  &\Checkmark&\textcolor{gray!50}{\XSolidBrush}&\Checkmark&\textcolor{gray!50}{\XSolidBrush}&+3.3&+3.9&\quad \textcolor{black}{-5.07}&+1.12\\
    \rowcolor{green!8}DETA++ w/ $\mathcal{L}_{clean}$ \& $\mathcal{L}_{noise}$  &\textcolor{gray!50}{\XSolidBrush}&\textcolor{gray!50}{\XSolidBrush}&\Checkmark&\Checkmark&+\textbf{3.7}&+\textbf{4.5}&\quad  \textcolor{black}{-\textbf{8.60}}& +\textbf{1.95}\\
    \hline
    \end{tabular} %
  \label{tablxxx}%
\end{table*}%

\subsubsection{Ablation Studies}
\label{oodablation}
\textcolor{black}{Section \ref{ablation} presents ablation studies that highlight the effectiveness of the devised components of DETA++ (i.e., CoRA, $\mathcal{L}_{clean}$, $\mathcal{L}_{noise}$, IntraSwap and LocalNCC) in enhancing the robustness of few-shot classification approaches to the dual noises in support samples.
As discussed  in Section \ref{fsood},  DETA++ only retains four components, i.e., CoRA, $\mathcal{L}_{clean}$, $\mathcal{L}_{noise}$ and  IntraSwap, to improve the noise robustness of existing prompt tuning based few-shot OOD detection methods  in the presence of noisy query samples.}
To understand the impacts of the four components, we conduct a component-wise analysis by alternatively adding one of them to the baseline {CoOp$_{\mathrm{MCM}}$ \cite{miyai2024locoop}} in Table \ref{abood}.
As can be seen, all the four components contribute to the overall performance improvement of DETA++.
\textcolor{black}{
It is also worth mentioning that the developed noise energy maximization loss $\mathcal{L}_{noise}$ establishes the most remarkable performance improvement among those components (by \underline{\textbf{4.18}}\%/8.6\% in FPR, \underline{\textbf{0.94}}\%/1.95\% in AUROC), which demonstrates the effectiveness of using the detected noisy regions of the CoRA module to improve the decision boundary between class features and noise features during the prompt tuning or task adaptation process.}

\subsection{Comparisons of the Losses in DETA and DETA++} \label{sect.losscompare}
\textcolor{black}{The DETA++ approach proposed in this work is an extension of DETA \cite{Zhang_2023_ICCV}. 
Beyond incorporating two new modules (i.e., IntraSwap and LocalNCC), the losses used in DETA++ differ significantly from those in DETA. \textcolor{black}{As illustrated in Fig.  \textcolor{magenta}{5}}, DETA develops a local compactness loss $\mathcal{L}_l$ and a global dispersion loss $\mathcal{L}_g$ to tackle image  and label noise in support samples. In contrast, DETA++ builds a clean prototype loss $\mathcal{L}_{clean}$ and a noise entropy maximization loss $\mathcal{L}_{noise}$ to address ID and OOD noises in both support and query samples. 
In particular, $\mathcal{L}_{clean}$ is an optimized version of $\mathcal{L}_g$, designed to pull clean regions closer to the their class prototypes constructed from clean images. $\mathcal{L}_{noise}$ is proposed to enhance the decision boundary between class features and noise features by maximizing the entropy of model predictions on ID and OOD noise. 
Quantitative comparisons of these losses are shown in Table \textcolor{magenta}{10}.
From the results in the table, we have the following observations.
\textbf{\underline{Firstly}}, integrated with $\mathcal{L}_{clean}$ and $\mathcal{L}_{noise}$, DETA++ achieves the best results across all settings, proving the advantages of $\mathcal{L}_{clean}$ and $\mathcal{L}_{noise}$ over the two losses $\mathcal{L}_l$ and $\mathcal{L}_g$ used in DETA.
\textbf{\underline{Secondly}}, replacing $\mathcal{L}_g$ with $\mathcal{L}_{clean}$ leads to greater performance improvements over the baselines, demonstrating the superiority of $\mathcal{L}_{clean}$ compared to $\mathcal{L}_g$ and the effectiveness of filtering out noisy features using the threshold $\varrho$.
\textbf{\underline{Thirdly}}, the devised $\mathcal{L}_{noise}$ of DETA delivers more significant improvements in few-shot OOD detection performance compared to other losses, underscoring its efficacy in enhancing the decision boundary between class features and ID/OOD noise features.
\textbf{\underline{Finally}}, improved classification results of ID classes consistently lead to enhanced few-shot OOD detection performance, aligning with previous findings in \cite{vaze2022openset,fort2021exploring}. 
In summary, DETA++'s $\mathcal{L}_{clean}$ and$\mathcal{L}_{noise}$ demonstrate greater effectiveness compared to DETA's $\mathcal{L}_{l}$ and$\mathcal{L}_{g}$.
}

\begin{figure} 
\setlength{\abovecaptionskip}{-0.3cm}  
	\centering
	\includegraphics[width=1\linewidth]{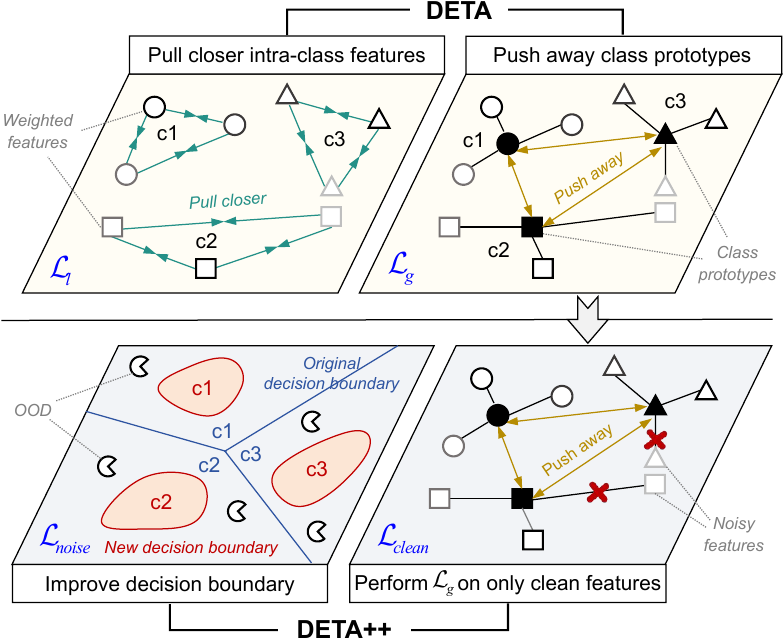}
        \textcolor{black}{
	\caption{Qualitative comparison of the losses in DETA \cite{Zhang_2023_ICCV} and DETA++.}}
		\label{fig.comparisons}  
\end{figure}

\begin{figure*}
\setlength{\abovecaptionskip}{0.cm}  
\setlength{\belowcaptionskip}{0.2cm} 
		\centering
		\includegraphics[width=1\linewidth]{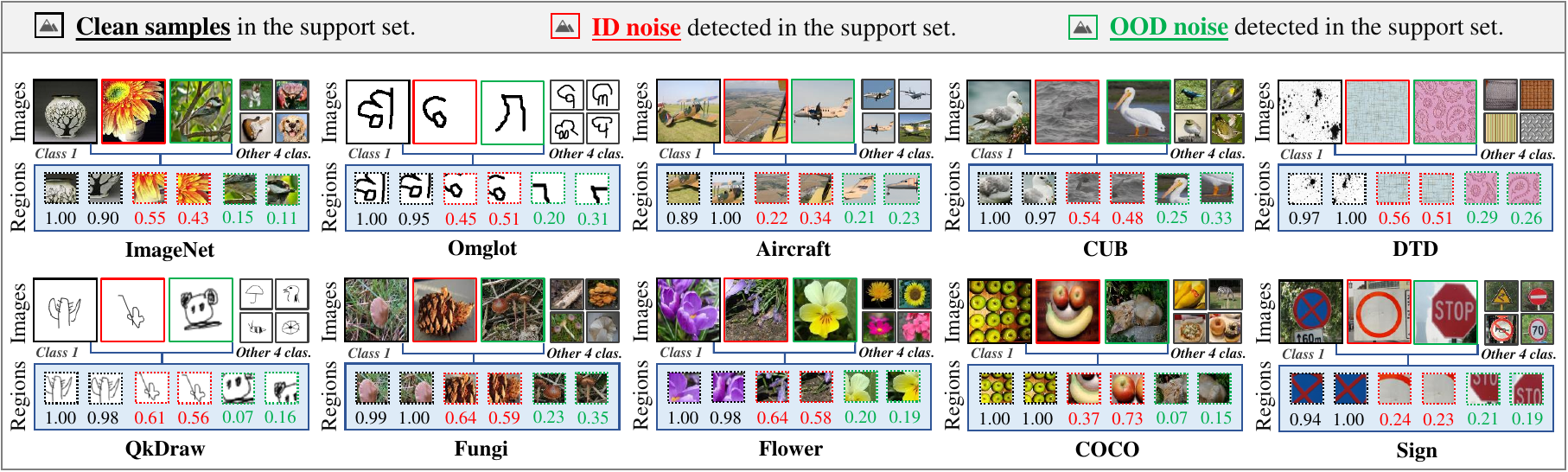} 
		\caption{Visualization of the cropped regions and calculated weights of the CoRA module for ten 5-way 10-shot tasks sampled from Meta-Dataset. 
        To facilitate comparison, the region weights for each class are divided by the maximum weight in the class. The baseline is full-FT\&MoCo.} 
		\label{vis1}
\end{figure*}

\subsection{Visualization}
\label{sectionvis}
Here, we provide visualization results to qualitatively investigate how the proposed DETA++ works. 
Fig. \ref{vis1} presents the visualization of the cropped regions as well as the calculated region weights by the CoRA module for ten 5-way 10-shot few-shot tasks constructed from the ten datasets of the Meta-Detaset benchmark, respectively.
As can be observed, the devised CoRA module successfully assigns larger (resp. smaller) weights to task-specific clean (resp. task-irrelevant noisy) image regions for each task. Hence, we can easily filter out ID and OOD noise in few-shot tasks to achieve noise-robust task adaptation and inference using the proposed DETA++ framework.

\section{Conclusions}
In this work, we reveal that the overlooked ID and OOD noise negatively affect the task adaptation performance of FSL methods. 
Motivated by this, we propose DETA++, which tackles the dual noises in a unified framework. 
Without extra supervision, DETA++ filters out task-irrelevant, noisy features from ID and OOD noise by taking advantage of the local region details of the support samples.
DETA++ is orthogonal to task adaptation based few-shot classification and OOD detection approaches,
thus can be used as a plugin to improve all of them.  
Extensive experiments demonstrate the effectiveness and flexibility of DETA++: DETA++ consistently improves competitive baselines and
establishes the state-of-the-art performance in both few-shot classification and OOD detection benchmarks. 

\textcolor{black}{
While our proposed DETA++ method offers several benefits, it primarily faces two limitations: 1) The additional hyperparameters may somewhat restrict the method's flexibility. 2) Although DETA++ has shown promising results in few-shot learning scenarios, its scalability to more complex tasks with significantly more support samples and classes requires further investigation. In future work, we aim to refine hyperparameter tuning processes and develop strategies to extend DETA++'s applicability to broader and more challenging scenarios.}


\section*{Acknowledgment}
This study is supported by grants from the Postdoctoral Fellowship Program of CPSF (Grant No. GZB20240625), the National Natural Science Foundation of China (Grant No. 62425208) and the Fundamental Research Funds for the Central Universities (Grant No. 2682025CX010),
the Science and Technology Innovation Committee of Shenzhen Municipality Foundation (Grant No. JCYJ20240813114208012).

\bibliographystyle{IEEEtran}
\bibliography{bare_adv}

%




\begin{IEEEbiography}[{\includegraphics[width=1in,height=1.25in,clip]{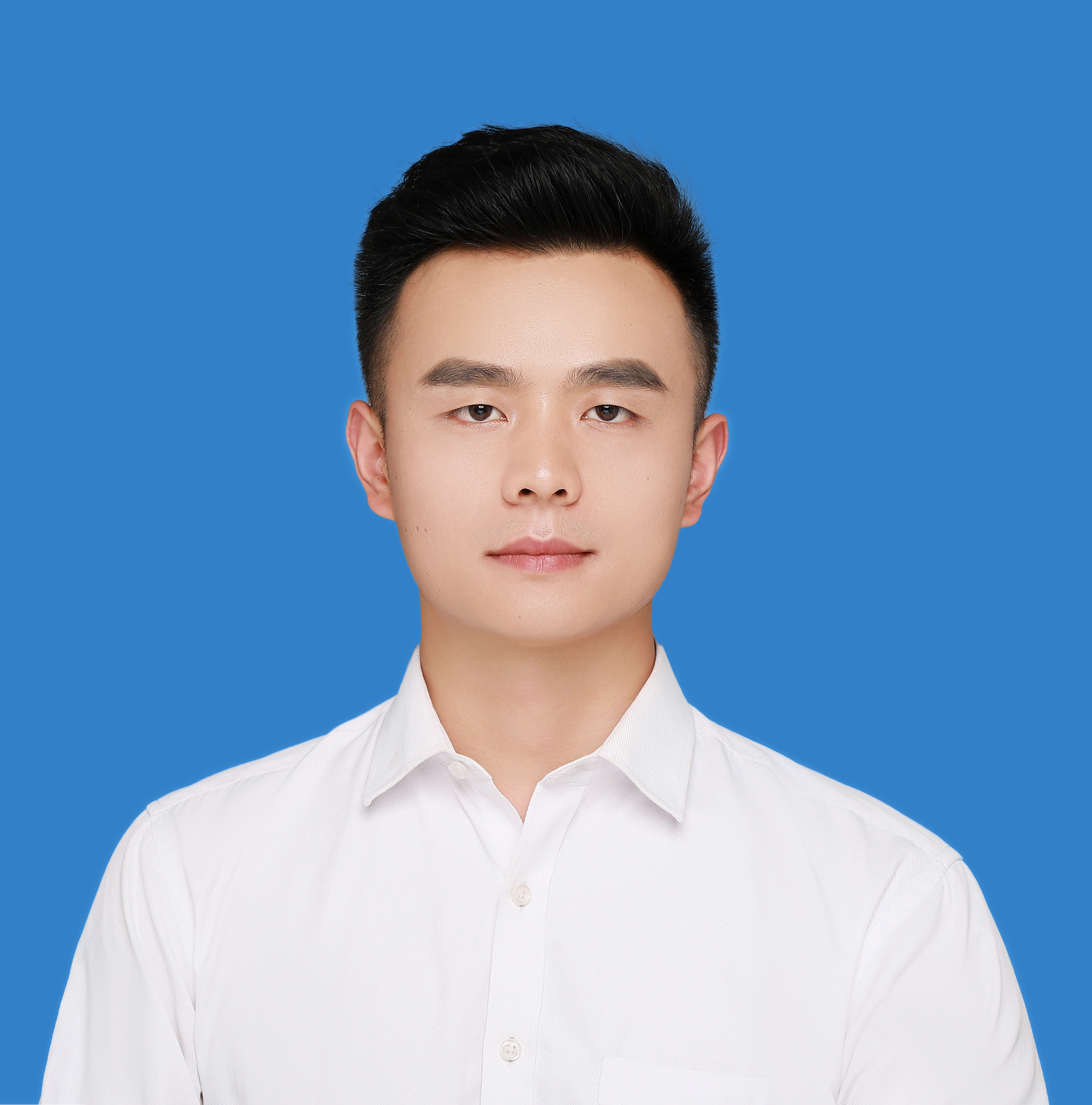}}]{Ji Zhang} is an Assistant Professor with the School of Computing and Artificial Intelligence, Southwest Jiaotong University, China. He obtained his PhD degree from University of Electronic Science and Technology of China in 2024, under the supervision of Prof. Jingkuan Song. His research interests include few-shot learning, transfer learning and robotics. 
He has published over ten papers on top conferences/journals, such as CVPR'24-25, ICCV'23, ICML'23, TIP'23. 
\end{IEEEbiography}

\begin{IEEEbiography}[{\includegraphics[width=1in,height=1.25in,clip]{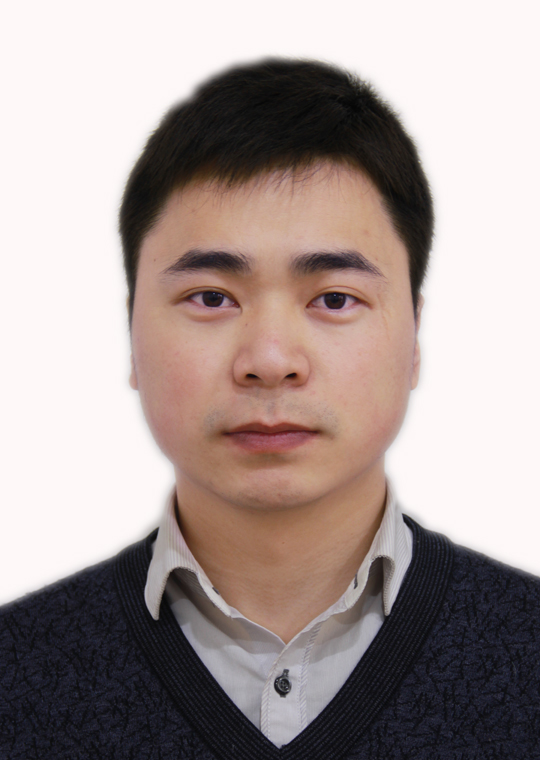}}]{Jingkuan Song} is a professor with the School of Computer
Science and Technology, Tongji University, China. 
He joined Columbia University as a Postdoc Research Scientist (2016-2017), and University of Trento as a Research Fellow (2014-2016). He obtained his PhD degree in 2014 from The University of Queensland (UQ), Australia. His research interest includes large-scale multimedia retrieval, LLMs and deep learning techniques. He was the winner of the Best Paper Award in ICPR (2016, Mexico), Best Student Paper Award in Australian Database Conference (2017, Australia), and Best Paper Honorable Mention Award (2017, Japan). He is an Associate Editor of IEEE TMM and ACM TOMM.
\end{IEEEbiography}

\begin{IEEEbiography}[{\includegraphics[width=1in,height=1.25in,clip]{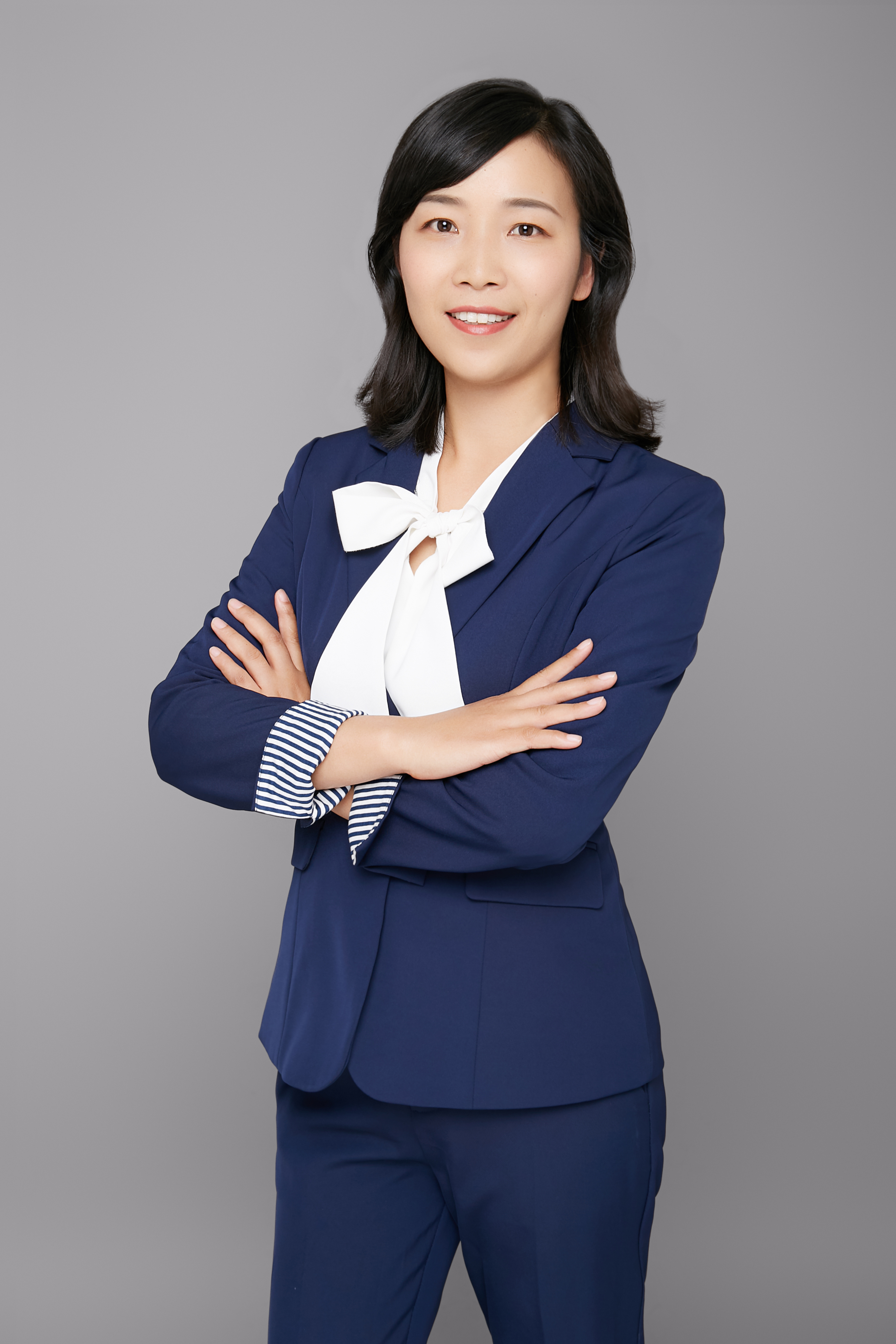}}]{Lianli Gao} is a professor with the School of Computer Science and Engineering, University of Electronic Science and Technology of China. She obtained her PhD degree in Information Technology from The University of Queensland (UQ), Australia, under the supervision of Prof. Jane Hunter and Prof. Michael Bruenig. Her research ranges from Semantic Web, Machine Learning, Deep Learning, Computer Vision (Images and Videos), NLP, Knowledge Reasoning, Knowledge and the related practical applications etc. Specifically, she is mainly focusing on integrating Natural Language for Visual Content Understanding. She has  the winner of the IEEE Transactions on Multimedia 2020 Prize Paper Award, Best Student Paper Award in Australian Database Conference (2017, Australia), IEEE TCMC Rising Star Award 2020 and ALIBABA Academic Young Fellow. She is an Associate Editor of IEEE TMM.
\end{IEEEbiography}

\begin{IEEEbiography}[{\includegraphics[width=1in,height=1.25in,clip]{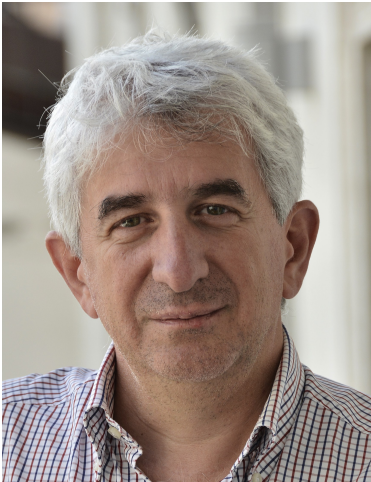}}]{Nicu Sebe} is a professor with the Department of Information Engineering and
Computer Science, University of Trento, leading the research in the areas of multimedia information retrieval and human behavior understanding. He was the General CoChair of ACM Multimedia 2013 and 2022, and the Program Chair of ACM Multimedia 2007 and 2011, ECCV 2016, ICCV 2017 and ICPR 2020. He is a fellow of the International Association for Pattern Recognition (IAPr) and of the European Laboratory for Learning and Intelligent Systems (ELLIS). 
\end{IEEEbiography}

\begin{IEEEbiography}[{\includegraphics[width=1in,height=1.25in,clip]{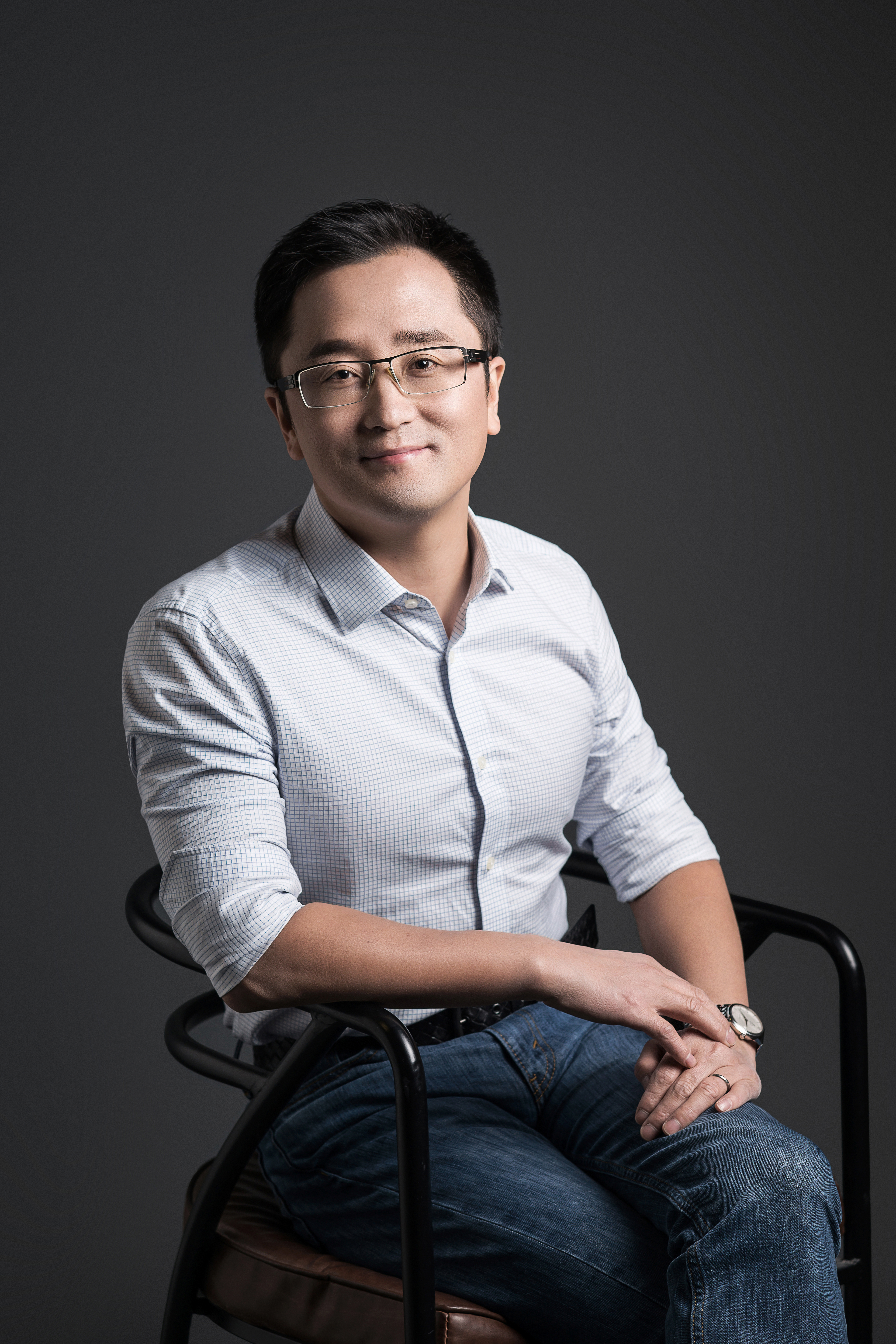}}]{Heng Tao Shen} is a professor with the School of Computer Science and Technology, Tongji University, China. 
He obtained his BSc with 1st class Honours and PhD from Department of Computer Science, National University  of  Singapore in 2000 and 2004 respectively.
His current research interests include multimedia search, computer vision, artificial intelligence, and big data management. He has published 300+ peer-reviewed papers and received 7 best paper awards from international conferences, including the Best Paper Award from ACM Multimedia 2017 and Best Paper Award-Honourable Mention from ACM SIGIR 2017. He has served as General Co-chair for ACM Multimedia 2021 and TPC Co-Chair for ACM Multimedia 2015, and is an Associate Editor of ACM Trans. of Data Science (TDS), IEEE Trans. on Image Processing (TIP), IEEE Trans. on Multimedia (TMM), and IEEE Trans. on Knowledge and Data Engineering (TKDE). He is a Fellow of ACM/IEEE/OSA.
\end{IEEEbiography}

\end{document}